\documentclass[10pt,twocolumn,letterpaper]{article}

\usepackage[pagenumbers]{cvpr} 

\usepackage{graphicx}
\usepackage{amsmath}
\usepackage{amssymb}
\usepackage{amsthm}
\newtheorem{definition}{Definition}[section]
\newtheorem{conjecture}{Conjecture}[section]
\usepackage{booktabs}

\usepackage{epsfig}
\usepackage[ruled, lined, linesnumbered, commentsnumbered, longend]{algorithm2e}

\usepackage[title]{appendix}

\newtheorem{theorem}{Theorem}[section]
\newtheorem{corollary}{Corollary}[theorem]

\newtheorem{innercustomthm}{Theorem}
\newenvironment{customthm}[1]
  {\renewcommand\theinnercustomthm{#1}\innercustomthm}
  {\endinnercustomthm}

\usepackage[pagebackref,breaklinks,colorlinks]{hyperref}

\usepackage[capitalize]{cleveref}
\crefname{section}{Sec.}{Secs.}
\Crefname{section}{Section}{Sections}
\Crefname{table}{Table}{Tables}
\crefname{table}{Tab.}{Tabs.}

\def\cvprPaperID{7248} 
\def\confName{CVPR}
\def\confYear{2022}

\begin{document}

\title{Maximum Consensus by Weighted Influences of Monotone Boolean Functions\thanks{This is the author version of a paper accepted in CVPR 2022.}}

\author{Erchuan Zhang$^*$ \quad David Suter$^*$ \quad Ruwan Tennakoon$^{**}$ \quad Tat-Jun Chin$^\dagger$ \\
 Alireza Bab-Hadiashar$^{**}$\quad Giang Truong$^*$ \quad Syed Zulqarnain Gilani$^*$\\
$^*$Edith Cowan University, Perth Australia\\
$^{**}$RMIT University, Melbourne Australia\\
$^\dagger$University of Adelaide, Adelaide Australia\\
{\tt\small {\{erchuan.zhang, d.suter, h.truong, s.gilani\}}@ecu.edu.au}, {\tt\small {\{ruwan.tennakoon,abh\}}@rmit.edu.au}\\
{\tt\small {tat-jun.chin}@adelaide.edu.au}
}

\maketitle

\begin{abstract}
Robust model fitting is a fundamental problem in computer vision: used to pre-process raw data in the presence of outliers. Maximisation of Consensus (MaxCon) is one of the most popular robust criteria and widely used. Recently (Tennakoon \etal.~CVPR2021), 
a connection has been made between MaxCon and estimation of influences of a Monotone Boolean function. 
Equipping the Boolean cube with different measures and adopting different sampling strategies (two sides of the same coin) can have differing effects:
which leads to the current study. This paper studies the concept of weighted influences for solving MaxCon.
In particular, we study endowing the Boolean cube with the Bernoulli measure and performing biased (as opposed to uniform) sampling. 
 Theoretically, we prove the weighted influences, under this measure, of points belonging to larger structures are smaller than those of points belonging to smaller structures in general.
 We also consider another ``natural'' family of sampling/weighting strategies, sampling with uniform measure concentrated on a particular (Hamming) level of the cube.
 
 Based on weighted sampling, we modify the algorithm of Tennakoon \etal, and test on both synthetic and real datasets. This paper is not promoting a new approach per se, but rather studying the issue of weighted sampling. Accordingly, we are not claiming to have produced a superior algorithm: rather we show some modest gains of Bernoulli sampling, and we illuminate some of the interactions between structure in data and weighted sampling.
\end{abstract}

\section{Introduction}
\label{sec:intro}

Robust model fitting is a fundamental problem in processing data in the presence of outliers, which is a long-standing and challenging research topic. 
{\em Maximisation of Consensus} (MaxCon) is one of the most popular criteria for robust fitting, which aims at finding a model that has the largest consensus on the given measurements. Formally, given a data set $\mathcal{X}=\{\pmb{x}_i\}_{i=1}^n$ of size $n$ and a prescribed threshold $\varepsilon>0$, the maximum consensus criterion for prescribed type of mathematical model is formulated as,
\begin{equation}\label{CM}
\begin{aligned}
\max_{\pmb{\theta}\in \Omega,\mathcal{I}\subseteq \mathcal{X}}~~~&|\mathcal{I}|,\\
s.t.~~~~~&r(\pmb{x}_i,\pmb{\theta})\leqslant \varepsilon,~~~\forall~\pmb{x}_i\in\mathcal{I},
\end{aligned}
\end{equation}
where $\Omega$ is a subset of $\mathbb{R}^p$, $p$ is the dimension of the prescribed mathematical model\footnote{In some cases, $p$ should be a dimension larger than the dimension of the prescribed model. For example, in rotation registration, $\pmb{\theta}\in SO(3)\subseteq \mathbb{R}^9$, $p=9$ rather than $dim(SO(3))=3$.}, $\pmb{\theta}$ is the model parameter, $|\mathcal{I}|$ is the size of the subset $\mathcal{I}\subseteq \mathcal{X}$, $r(\pmb{x}_i,\pmb{\theta})$ is the fitting residual of datum $x_i$ with respect to the model $\pmb{\theta}$.

Random sample consensus (RANSAC), ``the original MaxCon algorithm'', is probably one of the most influential and widely used method since it was proposed in the 1980s \cite{fischler1981random}. The key idea is to adopt a hypothesize and verification procedure, which  randomly samples subsets to fit the model and computes the consensus with respect to the obtained model. Such randomized methods can be very fast but there is no guarantee for the optimality of solutions. 
Variants of RANSAC, such as LO-RANSAC \cite{chum2003locally} and PROSAC \cite{chum2005matching}, may improve the original RANSAC, but, they still inherit the drawback of randomized methods.

Another type of approach is based on the $M$-estimator paradigm, or other ways to modify the residual measure. Examples include: $l_1$ method \cite{olsson2010outlier}, $l_\infty$ method \cite{sim2006removing,seo2009outlier}, iteratively reweighted least squares \cite{aftab2015convergence}, reweighted $l_1$ method \cite{purkait2017maximum}, exact penalty method \cite{le2017exact,le2019deterministic}. Such approaches  solve  MaxCon deterministically, but like RANSAC, only approximately.

It is known that the MaxCon problem is combinatorially hard \cite{chin2018robust}. Optimal methods,  
(branch-and-bound (BnB) search \cite{li2009consensus} and $A^*$ tree search \cite{chin2015efficient,cai2019consensus}) exist but their (worst case) runtime increases exponentially with the size of the problem, which makes them ineffective for large scale and high dimensional problems.

This work directly follows \cite{tennakoon2021consensus}
- here examining alternative definitions of influences and associated estimation methods.
That work, in turn, uses the same basic ``tree'' strategy of  Chin \etal  \cite{chin2015efficient}, which used $A^*$ search on a tree structure, founded on the notion of ``basis'' in LP-type theory. The  
$A^*$ search is able to find global optimal solutions. However, this method is slow in general, despite speedups introduced by 
Cai \etal  \cite{cai2019consensus}.
In \cite{truong2021unsupervised}, an unsupervised learning approach was proposed to determine which point in a basis to remove in solving robust model fitting problems. 
This approach adopted the framework of reinforcement learning, where removing points are guided by maximising {\em rewards}. Like many learning based approaches, it may take a long time to train and it is hard to analyse the method and its ability to generalise. 
Like \cite{truong2021unsupervised} and \cite{tennakoon2021consensus}, modifications to the basic tree search that lose the priority queue guarantees of $A^*$, sacrifice optimality guarantees for speed. This paper is also spiritually in that context.  
\subsection{Influence and consensus maximisation}

Recently, Tennakoon \etal~\cite{tennakoon2021consensus} characterized the maximum consensus problem by {\em Monotone Boolean Function Theory} and investigated the characterisation of the MaxCon solutions by {\em influences} 
of a monotone Boolean function. In detail, any subset $\mathcal{I}\subseteq \mathcal{X}=\{\pmb{x}_i\}_{i=1}^n$ can be represented by a bit vector $\pmb{b}$ of length $n$, where its $i$-th component $b_i=0$ denotes the exclusion of the datum $\pmb{x}_i$ and $b_i=1$ denotes the inclusion of the datum $\pmb{x}_i$. Then, any subset $\mathcal{I}$ is nothing more than a vertex of an $n$-dimensional Boolean cube. A Boolean function $f:\{0,1\}^n\rightarrow \{0,1\}$: 
\begin{align}\label{defmbf}
f(\pmb{b}):=\begin{cases}
1,~&\mathrm{if}~~r(\pmb{\theta},\pmb{x}_i)> \varepsilon\\
0,~&\mathrm{if}~~r(\pmb{\theta},\pmb{x}_i)\leqslant \varepsilon
\end{cases},~\forall~i\in\{j:b_j=1\}.
\end{align}
defines feasbility 
(the subset $\{\pmb{x}_i\}$ denoted by $\pmb{b}$ is called {\em feasible} if $f(\pmb{b})=0$ and {\em infeasible} otherwise). Solving MaxCon is a search for the feasible subset of maximal size.

\begin{figure}[t]
  \centering
   \includegraphics[width=0.7\linewidth]{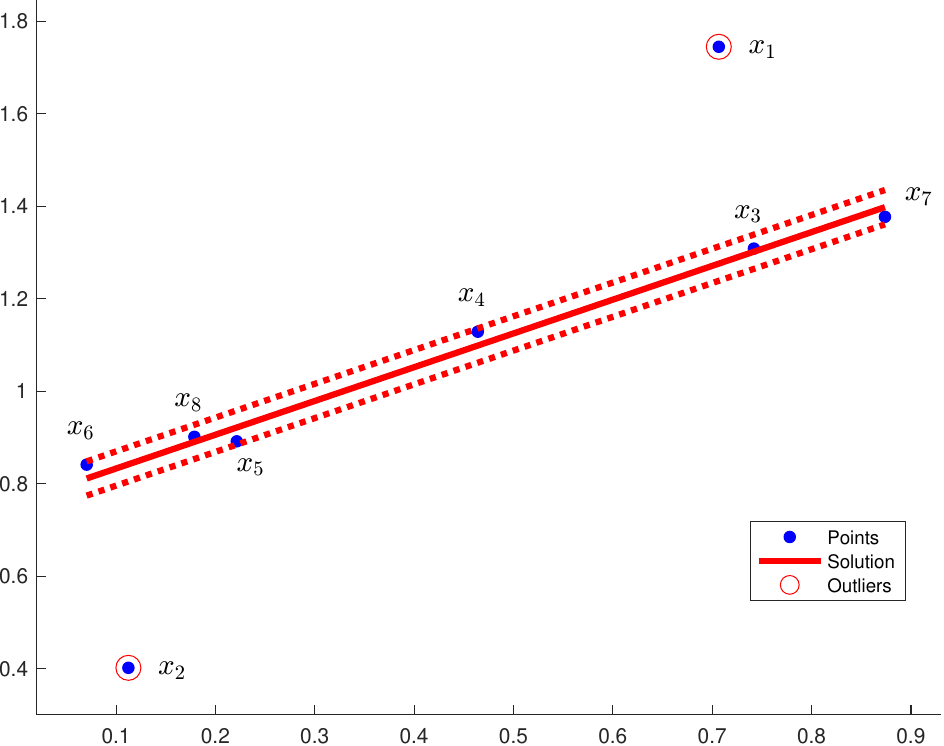}\\
   \includegraphics[width=0.7\linewidth]{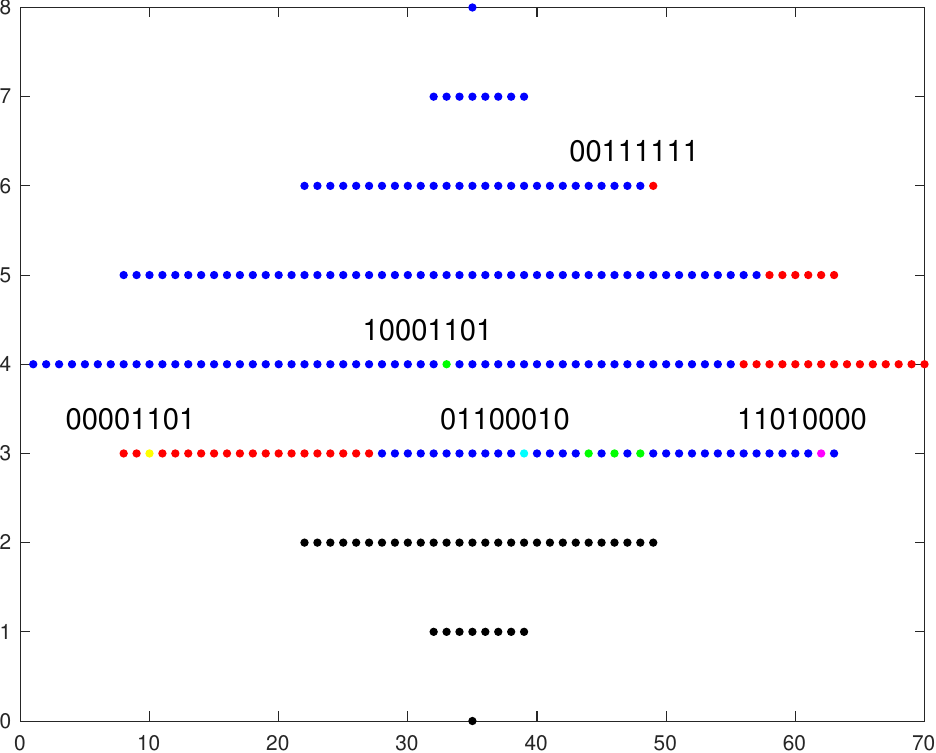}
   \caption{Top: 2D line fitting problem with $8$ points. Bottom: Boolean cube associated with the fitting problem. The $8$D Boolean cube is flattened on the plane, where every blue dot represents  an infeasible subset of the line fitting problem and other colored dots represent feasible subsets. These dots are arranged by the ordering relation $\prec$ from top to bottom. In this toy example, there are four upper zeros, \ie, $\pmb{b}^6=00111111$ (maximum upper zero, superscript represents how many data points are chosen), $\pmb{b}^4=10001101$, $\pmb{b}^3=011000010$, $\pmb{b}^{3^\prime}=11010000$, which are highlighted in the cube and the sub-cubes determined by these vertices are distinguished by different colors. The vertex $\pmb{b}^{3^{\prime\prime}}=00001101$ is a pseudo upper zero (see definition in Section \ref{sec:theory}). Note that all vertices below level $3$, \ie, less than $3$ points are fitted, must be feasible.}
   \label{fig:linecube}
   \vspace{-3mm}
\end{figure}

A Boolean function 
is called {\em monotone} if $f(\pmb{a})\leqslant f(\pmb{b})$ holds for any $\pmb{a}\prec \pmb{b}$, where the ordering relation $\prec$ means $i$-th component $a_i\leqslant b_i$ for any $1\leqslant i\leqslant n$. 
We can easily see that the Boolean function \eqref{defmbf} is monotone. The reason is that 
adding data points to infeasible subsets still keeps them \textit{infeasible} while deleting data points from feasible subsets keeps them \textit{feasible}. An {\em upper zero} $\pmb{b}$ of a monotone Boolean function $f$ is a vertex such that $f(\pmb{b})=0$ and $f(\pmb{a})=1$ for any $\pmb{a}$ satisfying $\pmb{b}\prec \pmb{a}$. A {\em maximum upper zero} $\pmb{b}$ of $f$ is an upper zero with the maximal size, \ie, $f(\pmb{b})=0$ and $f(\pmb{a})=1$ for any $\pmb{a}$ satisfying $\Vert \pmb{a}\Vert_1>\Vert \pmb{b}\Vert_1$, where $\Vert\cdot\Vert_1$ is the $l_1$ norm. The upper zeros essentially characterise all possible candidate consensus solutions systematically and finding maximum consensus is equivalent to finding the maximum upper zero of associated Boolean function. See Figure \ref{fig:linecube} for a toy example of $2$D line fitting, which illustrates the connection between monotone Boolean functions and the MaxCon problem.

The key concept in \cite{tennakoon2021consensus} is  {\em influence}. 
When endowing the Boolean cube with a uniform measure, the {\em influence} of $b_i$ of a Boolean function $f$ is defined \cite{o2014analysis} as,
\begin{align}\label{MBFinf}
{\rm{Inf}}_i [f] := \mathbb{P}_{\pmb{b}\sim \{0,1\}^n}\left[f(\pmb{b})\neq f(\pmb{b}^{\oplus i})\right],
\end{align}
where $\pmb{b}^{\oplus i}$ denotes flipping the $i$-th bit in $\pmb{b}$, $\pmb{b}\sim \{0,1\}^n$ means $\pmb{b}$ is uniformly distributed on the Boolean cube. 

To estimate the influences in equation \eqref{MBFinf}, the most obvious and natural way would be to uniformly sample, and use the empirical counts as (unbiased) estimates. The whole Boolean Cube is, unfortunately, so vast, that spreading the samples uniformly is not an efficient sampling strategy. This undoubtedly led to the ``uniform at level $p'$'' (for $p'$ slightly larger than the combinatorial dimension $p$)  sampling strategy of  \cite{tennakoon2021consensus}, which will produce {\em biased estimates}. 

\subsection{Motivation of this paper}
Given the above, one can thus view the method of \cite{tennakoon2021consensus} as employing {\em biased} estimates of {\em uniform measure} influence (equation \eqref{MBFinf}). Alternatively, one could take the sampling distribution used in 
\cite{tennakoon2021consensus} as implicitly defining {\em a new (weighted) influence measure} (and the sampling strategy is now an unbiased estimator of this new influence measure). Both views are really two sides of the same coin. However, 
the primary issue is whether the biased estimates of uniform measure influence (first viewpoint: biased estimator - second viewpoint: new influence measure) actually preserve the ordering of the sizes of influences.

Put simply,   
in \cite{tennakoon2021consensus} there is a disconnect between the measure used in the definition (and also in the proof of influence being greater for outliers) and the measure used in the empirical estimation: but one can reconcile this by using a definition of influences that entails the same biased measure as would be used in sampling (to estimate the influences). We are thus led to a more systematic and coherent study of a topic called ``weighted'' (or biased) influences.

In section \ref{subsec:weightinf}, we equip the Boolean cube $\{0,1\}^n$ with the Bernoulli(q) measure $\mu_q$, namely, the probability of $b_i=1$ is $q$ and the probability of $b_i=0$ is $1-q$, which includes the uniform measure as a special case  ($q=\frac{1}{2}$). 
For this, an unbiased estimator is:
\begin{align}\label{oldinf}
{\rm{Inf}}_i^{(q)}[f] = \underset{\pmb{x}\sim \mu_q(\pmb{x})}{\mathbb{E}}[\mathbb{I}(f(\pmb{x})\neq f(\pmb{x}^{\oplus i}))],
\end{align}
where $\mathbb{I}(\cdot)$ is the indicator function, $\mathbb{E}[\cdot]$ is the sample mean of random variables.
Empirical evidence (see section \ref{sec:experiments}) shows that a Bernoulli weighted influence based method can achieve similar optimal solutions with less runtime or even better results with similar runtime. But perhaps more importantly, we analytically prove that this weighted measure preserves the necessary ordering of influences  - see section \ref{sec:theory}. In fact, since the expressions we derive for Bernoulli weighted influence subsume uniform influence as a special case, and since we cover more general data characteristics (section \ref{subsec:nonidealBern}) than that in the corresponding sections of \cite{tennakoon2021consensus}, this paper provides a more comprehensive theoretical analysis of the uniform measure as well.

Another family of measures, we call Hamming(k), puts equal mass on subsets of a fixed size (fixed Hamming norm) k, and zero elsewhere.
A particular case of which is applied in \cite{tennakoon2021consensus}. 
For these measures, it turns out that we do not need to conduct an analysis from scratch. The expressions we have for Bernoulli(q) weighted influence naturally stratify by level (as do of course the corresponding equations in \cite{tennakoon2021consensus} which are the special case Bernoulli(0.5)). This allows us to obtain the number of feasibility/infeasibility flips per level by inspection (and the corresponding weighted influence is then just the appropriate normalisation produced by dividing these counts by the number of vertices/subsets in the given level). Nonetheless, this was not pointed out in \cite{tennakoon2021consensus}, nor were any resulting observations made. Moreover, since we analyse a more general set of data assumptions 
we also have a more comprehensive picture than can be deduced simply by inspection of equations in \cite{tennakoon2021consensus}. 

Our main contributions can be summarised as:
\begin{itemize}
\item We characterize the influence of outliers/inliers in MaxCon by the concept of weighted influence in the Boolean function theory. By analysis, we prove that the weighted influences of points belonging to the larger structure in the data are generally smaller for both ideal and non-ideal cases (see section \ref{sec:theory} for definitions). Note that our analysis is much more general, and correctly aligned with sampling, compared to \cite{tennakoon2021consensus}.
Albeit, from the very large set of weighted influences one could define, we examine only two families of weighted influence: Bernoulli(q) and Hamming(k). 
\item We empirically test modifications made to the basic algorithm of \cite{tennakoon2021consensus}, to accommodate weighted sampling by Bernoulli measure. 
On several robust fitting tasks, our weighted influence based variant is an effective alternative to the version in \cite{tennakoon2021consensus}.
\end{itemize}
We also provide a conjecture, motivated by the structure of the equations we derive, that possibly links and explains the inherent structural complexity of MaxCon, by some recent constructs of what characterises hard problems - see Conjecture \ref{conjectureOGP} section \ref{subsec:nonidealBern}.

In concluding this introduction we emphasize that we are not proposing a hugely novel and vastly superior algorithmic approach. The aims and claims in the paper are in relation to investigating MaxCon under new definitions of influence, and corresponding estimation strategies. The spirit is a more careful and comprehensive study of those issues - not of producing the ``next leaderboard winning'' algorithm. 
\subsection{Notation}

We adopt the following notation: (1) Bold $\pmb{b}$ denotes a vertex (a vector of $n$ bits) in the Boolean cube $\{0,1\}^n$ and its $i$-th component is denoted by $b_i$. (2) $[n]=\{1,2,\cdots,n\}$. (3) $L_k:=\{\pmb{b}\in\{0,1\}^n|~ \Vert \pmb{b}\Vert_1=k\}$ denotes level $k$ of the Boolean cube $\{0,1\}^n$ for $0\leqslant k\leqslant n$. (4) $L_{\leqslant k}:=\{\pmb{b}\in\{0,1\}^n|~ \Vert \pmb{b}\Vert_1\leqslant k\}$ represents levels below level $k+1$ in the cube $\{0,1\}^n$. Similar definitions for $L_{<k}, L_{\geqslant k}, L_{>k}$. (5) For $\pmb{b^*}\in L_k$, $B_{\pmb{b^*}}:=\{\pmb{b}\in \{0,1\}^n | \Delta(\pmb{b},\pmb{b^*})=l,\pmb{b}\in L_{k-l},1\leqslant l\leqslant k-p-1\}$ denotes the sub-cube determined by $\pmb{b^*}$, where $\Delta(\pmb{b},\pmb{b^*})=|\{i:b_i\neq b^*_i\}|$ is the Hamming distance between $\pmb{b}$ and $\pmb{b^*}$. (6) $S^j_{\pmb{b}^{k_i}}:=\{l\in [n] ~|~ b^{k_i}_l=j\}$, where $j=0,1$. $S^1_{\pmb{b}^{k_i}}$ and $S^0_{\pmb{b}^{k_i}}$ represent the set of indices of data points who are inliers and outliers with respect to $\pmb{b}^{k_i}\in L_{k_i}$, respectively. 

\section{Weighted Influences}\label{foundation}

In this section, we will introduce the definitions of (Bernoulli) weighted influences (section \ref{subsec:weightinf}) 
of monotone Boolean functions in section  and (Uniform Hamming) weighted influences (section  \ref{subsec:weightinfHamming}). 

\subsection{Bernoulli(q) weighted influences}\label{subsec:weightinf}

Let $\Omega_n=\{0,1\}^n$ be the discrete $n$-dimensional cube endowed with the Bernoulli measure $\mu_q$, where $\mu(\{\pmb{b}:b_i=1\})=q\in (0,1)$, namely, $\mu_q(\pmb{b})=q^{\Vert \pmb{b}\Vert_1}(1-q)^{n-\Vert \pmb{b}\Vert_1}$. Then, we can define a metric on $\Omega_n$ \cite{o2014analysis},
\begin{equation*}
\langle f_1,f_2\rangle~:=~\sum_{\pmb{b}\sim \Omega_n}f_1(\pmb{b})f_2(\pmb{b})\mu_q(\pmb{b})
\end{equation*}
for any $f_1,f_2$ defined on $\Omega_n$. The parity functions
\begin{equation*}
\chi_S^q(\pmb{b}):=q_{-}^{\sum_{i\in S}b_i}\cdot q_{+}^{|S|-\sum_{i\in S}b_i}
\end{equation*}
form a basis of the space $(\Omega_n,\langle\cdot,\cdot\rangle)$, where $S$ is a subset of $[n]$, $\pmb{b}\in \Omega_n$, $q_{-}=-\sqrt{\frac{1-q}{q}}$, $q_{+}=\sqrt{\frac{q}{1-q}}$. $\chi_{\emptyset}^q(\pmb{b})\equiv 1$ by convention.

The {\em weighted first-order Fourier coefficient} of a Boolean function $f$ on $i$-th variable can be given by
\begin{align}\label{fouriercoe2}
\hat{f}^q(\{i\})=\langle f,\chi_{\{i\}}^q\rangle=\sum_{\pmb{b}\sim \Omega_n}f(\pmb{b})q_{-}^{b_i}q_{+}^{1-b_i}\mu_q(\pmb{b}).
\end{align}

With respect to the Bernoulli measure $\mu_q$, the {\em weighted influence} of the $i$-th variable on a Boolean function $f$ defined on $\Omega_n$ is defined as (see Page 9 in \cite{kalaithreshold})
\begin{align}\label{weightinfdef}
{\rm{Inf}}^q_i[f]:=\mu_q(\{\pmb{b}:f(\pmb{b})\neq f(\pmb{b}^{\oplus i})\}).
\end{align}

The relationship between $\hat{f}^q(\{i\})$ and ${\rm{Inf}}^q_i[f]$ is given by the following theorem\footnote{Refer to Appendix \ref{sec:proofs} for the proof. Similar result can be found in \cite{kalaithreshold} (Page 20) but without proof.}.

\begin{theorem}\label{inffour2}
If $f:\{0,1\}^n\rightarrow \{0,1\}$ is a monotone Boolean function, then ${\rm{Inf}}_i^q[f]=-\frac{1}{\sqrt{q(1-q)}}\hat{f}^q(\{i\})$.
\end{theorem}

In practice, it is not very realistic to calculate the exact value of \eqref{fouriercoe2} or \eqref{weightinfdef} since the summation or probability contains $2^n$ possible $\pmb{b}$ to evaluate. Since we will take advantage of the order information of all weighted influences rather than their exact values to design algorithms for MaxCon, 
it is sufficient to get approximated weighted influences of good quality. 

\begin{figure}[ht]
  \centering
   \includegraphics[width=0.5\linewidth]{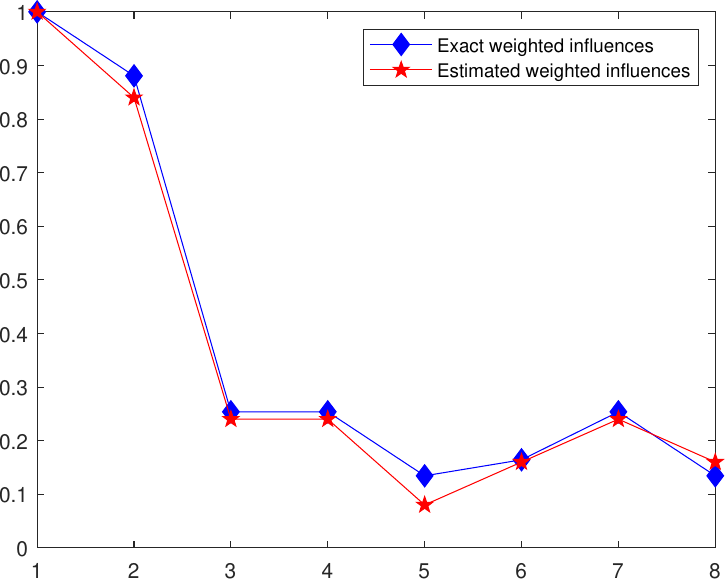}
   \caption{Comparison of normalized exact and estimated weighted influences of all data points used in Figure \ref{fig:linecube}, where $q$ is set to $0.5$ and we set the number of samples $h=100$.}
   \label{fig:lineextestinf}
   \vspace{-3mm}
\end{figure}

The way we estimate weighted influences is illustrated as follows: Given a sample size $h>p$ and a weight $q\in (0,1)$, we sample half of a set of vertices $\{\pmb{b}_j\}_{j=1}^{\lfloor \frac{h}{2}\rfloor}$ on the Boolean cube $\Omega_n$ according to the Bernoulli measure $\mu_q$ and then flip $i$-th bit in all $\pmb{b}_j$ to generate another half samples $\{\pmb{b}_j\}_{j=\lfloor \frac{h}{2}\rfloor+1}^{h}$. The monotone Boolean function $f$ is evaluated on these vertices. Therefore, by \eqref{fouriercoe2}, \eqref{weightinfdef} and Theorem \ref{inffour2}, we estimate the weight influence $\widetilde{{\rm{Inf}}}_i^q[f]$ as
\begin{small}
\begin{align}\label{estweightinf}
\widetilde{{\rm{Inf}}}_i^q[f]=-\frac{1}{h\sqrt{q(1-q)}}\sum_{j=1}^h f(\pmb{b}_j)q_{-}^{b_{j,i}}q_{+}^{1-b_{j,i}}\mu_q(\pmb{b}_j),
\end{align}
\end{small}
where $b_{j,i}$ is the $i$-th component of $\pmb{b}_j$. Note that the evaluate of $f$ on $\pmb{b}_j$ can be simplified by the monotonicity of $f$: if $f(\pmb{b}_j)=0$ and $b_{j,i}=1$, then $f(\pmb{b}_j^{\oplus i})=0$; if $f(\pmb{b}_j)=1$ and $b_{j,i}=0$, then $f(\pmb{b}_j^{\oplus i})=1$. Figure \ref{fig:lineextestinf} shows the estimated and exact weighted influences of data points illustrated in Figure \ref{fig:linecube}, where influences are normalized by maximal influence.

\subsection{Hamming(k) weighted influences}\label{subsec:weightinfHamming}

Defining a Fourier transform (and in general analysis) restricted to the Boolean slice  (level of equal Hamming norm) is actually a more complex topic, only recently researched. See for example,
\cite{wimmer2014low,Filmus2016a,Filmus2016b} - arguments that might appeal to Fourier theory for the uniform cube, often side-step the more complicated Fourier picture on the slice. Likewise, we will make no reference to Fourier coefficients.

In essence, one can go ``direct'' from the definition of expectations of the counts of transitions (from feasibility to infeasibility) by flipping the $i$-th bit (including or excluding the $i$-th data point) to the appropriate definition of influence for the slice-concentrated measures (and thereby for sampling/estimation strategies). Moreover, since these transitions are counted by level (slice) in the appropriate formulae for the Bernoulli(q) influences (given in section \ref{sec:theory} or in \cite{tennakoon2021consensus} for the Bernoulli(0.5) measure): one does not have to repeat complicated derivations/analysis but can merely ``pick off'' the relevant counts for the level under consideration, and then normalize by the total count measure on a slice $C^N_k$ for slice k. Moreover, since one is only interested in relative sizes of influences, normalization can be omitted. 

\section{Analysis of weighted influence}\label{sec:theory}

In this section, we theoretically show that points belonging to larger structures have smaller weighted influences: intuitively, the weighted influence of an ``inlier" data point is smaller than that of an ``outlier" data point. 
Proofs of all theoretical results in the following two subsections can be found in Appendix \ref{sec:proofs}. Informed by our analysis, once can search for the MaxCon solution, using weighted sampling for influence: with some assurance the estimates have the right relative order to identify outliers. 

\subsection{Bernoulli(q) - Ideal case}\label{subsec:idealcase}

Real datasets are ``non-ideal", \ie, there are multiple structures in the dataset and these often share many data points; yet the ideal case is a useful starting point.
\begin{definition}[Ideal Structure]
Suppose $\{\pmb{b}^{k_i}\}_{i=1}^{n_0}$ are upper zeros of a monotone Boolean function, where $\pmb{b}^{k_i}\in L_{k_i}$, $p+1\leqslant k_1\leqslant k_2\leqslant \cdots\leqslant k_{n_0}\leqslant n$. Then

$f$ is called {\em ideal} if $\forall ~i,j\in [n_0]$, $|S_{\pmb{b}^{k_i}}^1\cap S_{\pmb{b}^{k_j}}^1|\leqslant p$, namely, 
the structures have very small overlap.

$f$ is called {\em non-ideal} if $\exists~i,j\in [n_0]$, $|S_{\pmb{b}^{k_i}}^1\cap S_{\pmb{b}^{k_j}}^1|> p$. Namely, at least two structures have significant overlap. We call $\pmb{b}^{\alpha}=(b_1^{\alpha},\cdots,b_n^{\alpha})$ defined by
\begin{small}
\begin{equation*}
\begin{aligned}
b_l^{\alpha}:=\begin{cases}
1, ~~&l\in S_{\pmb{b}^{k_i}}^1\cap S_{\pmb{b}^{k_j}}^1,\\
0,&\mathrm{otherwise},
\end{cases}~~\forall~l\in [n],
\end{aligned}
\end{equation*}
\end{small}
a {\em pseudo upper zero} with respect to $\pmb{b}^{k_i}$ and $\pmb{b}^{k_j}$, where $\alpha:=|S_{\pmb{b}^{k_i}}^1\cap S_{\pmb{b}^{k_j}}^1|$ is the level that the pseudo upper zero belongs to. Suppose $\{\pmb{b}^{\alpha_i}\}_{i=1}^{m_0}$ $(p+1\leqslant \alpha_1\leqslant \alpha_2\leqslant \cdots\leqslant \alpha_{m_0}<k_{n_0})$ is the set of all pseudo upper zeros if $f$ is non-ideal.
\end{definition}

\begin{theorem}\label{idealsingle}({\bf Ideal Single Structure})
If $n_0=1$, namely, $f$ is ideal with respect to a single maximum upper zero $\pmb{b}^{k_1}$, then, for $i\in S_{\pmb{b}^{k_1}}^1$ (inliers),
\begin{small}
\begin{equation}
\label{eqn:inlier_ideal}
{\rm{Inf}}_i^q[f]=
(C^{n-1}_p-C^{{k_1}-1}_p)q^p(1-q)^{n-p-1},
\end{equation}
\end{small}
and for $i\in S_{\pmb{b}^{k_1}}^0$ (outliers),
\begin{small}
\begin{equation}
\label{eqn:outlier_ideal}
{\rm{Inf}}_i^q[f]=C^{n-1}_pq^p(1-q)^{n-p-1}+\sum_{l=p+1}^{k_1}C^{k_1}_lq^l(1-q)^{n-l-1},
\end{equation}
\end{small}
which implies ${\rm{Inf}}_{i_1}^q[f]-{\rm{Inf}}_{i_2}^q[f]=C^{{k_1}-1}_pq^p(1-q)^{n-p-1}+\sum_{l=p+1}^{k_1}C^{k_1}_lq^l(1-q)^{n-l-1}>0$, where $i_1\in S_{\pmb{b}^{k_1}}^0$, $i_2\in S_{\pmb{b}^{k_1}}^1$.
\end{theorem}

This theorem indicates that if there is only one ideal structure, namely, points are either inliers or outliers with respect to that structure, the weighted influences of outliers (all outliers share the same weighted influences) are strictly larger than those of inliers (all inliers share the same weighted influences).


\begin{theorem}\label{idealkthm}({\bf Ideal Multiple Structure})
If $n_0>1$, \ie, $f$ is ideal with several upper zeros, then, $\forall~i\in \cap_{l=1}^{n_0}S_{\pmb{b}^{k_i}}^{i_l}$ (if non-empty), ${\rm{Inf}}_i^q[f]$ is
\begin{small}
\begin{equation}\label{idealk}
\begin{aligned}
&(C^{n-1}_p-\sum_{i_s=1}C^{k_s-1}_p)q^p(1-q)^{n-p-1}\\
&+\sum_{i_s=0}\sum_{l=p+1}^{k_s}C^{k_s}_lq^l(1-q)^{n-l-1}.
\end{aligned}
\end{equation}
\end{small}
\end{theorem}
Put simply, we can see (from the summations over $i_s$) that the influence of a data point decreases when belonging to a structure (and the decrease is more the larger the structure is) and increases for every structure the point does not belong to (again, by more if that structure is large).
These increases and decreases are weighted differently (terms involving $q$), complicating the relationship.

Note: with multiple structures, one has to be careful to qualify ``inlier'' and ``outlier''. These only have meaning with respect to a nominated single structure. For that structure, all points ``inlier'' to other structures are actually outlier to the nominated one. 


From \eqref{idealk}, we find that the influence ${\rm{Inf}}_i^q[f]$ of data point $i$ is larger if it is an outlier with respect to more upper zeros. (So of course data outlier to all structures will have the largest influence of all).

Here, we introduce a new Boolean cube $\{0,1\}^{n_0}$, for any $i\in\cap_{l=1}^{n_0}S_{\pmb{b}^{k_i}}^{i_l}$, it corresponds to a vertex $(i_1,i_2,\cdots,i_{n_0})\in \{0,1\}^{n_0}$. Then, ${\rm{Inf}}_i^q[f]$ is a real-valued Boolean function on $\{0,1\}^{n_0}$. To shorten notation, we denote ${\rm{Inf}}_i^q[f]$ for $i\in \cap_{l=1}^{n_0}S_{\pmb{b}^{k_i}}^{i_l}$ by $\tilde{f}^q(S^{\pmb{i}})$.

\begin{corollary}\label{idealordercor}
The influences \eqref{idealk} have the following order relationship
\begin{align}
\forall~\pmb{i},\pmb{j}\in\{0,1\}^{n_0},~\pmb{i}\succ \pmb{j}\implies \tilde{f}^q(S^{\pmb{i}})<\tilde{f}^q(S^{\pmb{j}}).
\end{align}
Note that $\tilde{f}^q(S^\bullet)$ exists only if $S^\bullet$ is non-empty.
\end{corollary}


\subsection{Bernoulli(q) - Non-ideal case}
 \label{subsec:nonidealBern}
Given a pseudo upper zero $\pmb{b}^{\alpha_i}$, we introduce the notation
$
S_{\pmb{b}^{\alpha_i}}^j=\{l\in [n]| b_l^{\alpha_i}=1-j\},
$
where $j=0,1$. For any $i\in (\cap_{l=1}^{n_0}S_{\pmb{b}^{k_l}}^{i_l})\cap(\cap_{l=1}^{m_0}S_{\pmb{b}^{\alpha_{n_0+l}}}^{i_l})$ (if non-empty), denote the influence ${\rm{Inf}}_i^q[f]$ of $i$-th variable on $f$ by $\tilde{f}^q(S^{\pmb{i}})$, where $\pmb{i}=(i_1,i_2,\cdots, i_{n_0+m_0})$.  

Observe that if $\pmb{b}^{\alpha_{i_0}}$ is a pseudo upper zero with respect to upper zeros $\pmb{b}^{k_{i_1}}$ and $\pmb{b}^{k_{i_2}}$, a data point $i$ is an inlier with respect to $\pmb{b}^{k_{i_1}}$ and $\pmb{b}^{k_{i_2}}$, then it must be an inlier with respect to $\pmb{b}^{\alpha_{i_0}}$; if $i$ is an outlier with respect to any upper zero, then it must be an outlier with respect to $\pmb{b}^{\alpha_{i_0}}$.

\begin{theorem}\label{nonidealthm}
If $f$ is non-ideal, then the weighted influence $\tilde{f}^q(S^{\pmb{i}})$ is given by
\begin{small}
\begin{equation*}\label{nonidealeq}
\begin{aligned}
& (C^{n-1}_p-\sum_{\underset{1\leqslant s\leqslant n_0}{i_s=1}}C^{k_s-1}_p+\sum_{\underset{1\leqslant s\leqslant m_0}{i_{n_0+s}=0}}C^{\alpha_s-1}_p)q^p(1-q)^{n-p-1}\\
&+(\sum_{\underset{1\leqslant s\leqslant n_0}{i_s=0}}\sum_{l=p+1}^{k_s}C^{k_s}_l-\sum_{\underset{1\leqslant s\leqslant m_0}{i_{n_0+s}=1}}\sum_{l=p+1}^{\alpha_s}C^{\alpha_s}_l)q^l (1-q)^{n-l-1}.
\end{aligned}
\end{equation*}
\end{small}
where $\tilde{f}^q(S^\bullet)$ doesn't exist if $S^\bullet=\emptyset$.
\end{theorem}

The core strategy to prove this theorem is by induction on the number of pseudo upper zeros and taking advantage of Theorem \ref{idealkthm}. Similar to the ideal case, by Theorem \ref{nonidealthm}, we can see that the weighted influences of points belonging to larger structures are smaller. 
What becomes apparent is that with many structures, the expressions become hugely complicated by the combinatorics of possible intersections/overlap between them - complicating the measure of inlier/outlier dichotomy (with respect to any given structure). This leads us to conjecture:
\begin{conjecture}\label{conjectureOGP} {\bf (Complexity of MaxCon and Overlap Gap Property (OGP))}
In \cite{Gamarnike2108492118} (and the works referred therein), the computational complexity of many algorithms has been linked to what is called the OGP. Essentially it is argued that if the solution space is highly clustered (and here locality is measured by overlap of the solutions) then the problem will be hard for whole classes of algorithms. In short, the existence of potential solutions that are either very similar/have large overlap, or are widely separately (few intermediate separated) - is a ``signature'' of a hard  problem to solve. We conjecture that this is true of the MaxCon landscape of possible data configurations, and the above reflects (through the lens of influence) how this complicated data instance scenario manifests in a property related to the sought solution - influence. 
\end{conjecture}
This conjecture is particularly intriguing  because of the link between maximum independent set (studied in \cite{Gamarnike2108492118}) and MaxCon - it can be shown that the MaxCon solution is the maximum independent set of the (hyper)graph formed by the infeasible minimal sized subsets (and also to the minimum vertex cover of the complement hypergraph).  

\subsection{Uniform Hamming measure Influence}
\label{subsec:Hamming}

Consider the ideal single structure case
define in Theorem \ref{idealsingle}. 
Firstly, it is easy to see the for the ideal single structure, any algorithm that starts with a feasible set of size large than the combinatorial dimension, and then greedily adds points if the larger set remains feasible, will obtain the MaxCon solution. So it is an easy problem with obvious solution strategies. But if we did decide to use influences we can note that
from equation \eqref{eqn:inlier_ideal} the inlier influences came from counting feasible/infeasible transitions between levels $p$ and $p+1$ only. Thus for sampling uniform Hamming level above level $p+1$, there are no feasibility/infeasibility transitions caused by inliers. In other words, the Uniform Hamming influence measure, at that level or above, will be exactly zero for inliers. Since the influences of outliers, for the same measure will be non-zero, this seems to promise a remarkably efficient sampling strategy - one could eliminate outliers at the first sample that revealed a count for the associated influence - without the need to continue with the full estimation process.  Of course, practically, this is too good to hold for real data - it is very brittle to our strict assumptions here. Nonetheless, it does hint at the usefulness of a less brittle strategy of early termination of the counting process once a count reaches some degree of statistically significantly higher than the rest, rather than a set number of samples always being used: a future research topic. 

Now consider the ideal multiple structure setting. 
From equation \ref{idealk},  the influence accrued by being inlier to some structure (first term) is only accrued between level $p$ and $p+1$. The subtraction is due to feasibility transitions ``that didn't occur'' because the subset with added inlier remains within the same structure). So once again, sampling above level $p$ will not ``see'' those counts. But since inliers to one structure are outliers to another (we assumed no significant overlap): hence the influence of inliers to any structure will not be zero - different to the single structure case, as all structures are outliers to to (all) other structures and thus accrue influence from the second term in the equation. It is also easy to see that so long as the level is ``not too high'' (above the largest structure) the Hamming sampled influences will be an appropriate guide (influence of inliers of a larger structure will be smaller). (In that second term the largest structure is excluded from the sum over $i_s=0$, when calculating influence for that structure.) 

For space reasons, we relegate  further discussion to Appendix \ref{sec:hamming}.



\section{Experiments with Bernoulli measure}\label{sec:experiments}

In this section, we empirically demonstrate that Bernoulli weighted  influence behaves generally as our analysis predicts (essentially retaining the correct ordering of influences for inlier/outlier separation) across  several robust model fitting tasks on both synthetic and real datasets.The main algorithm we use is presented in Appendix \ref{sec:alg} (Algorithm 1 - which follows \cite{tennakoon2021consensus} ). To reduce the risk of losing genuine inliers, we also implement the local expansion (see Algorithm 2 in Appendix \ref{sec:alg}) at the last step in Algorithm 1, which likewise follows the strategy of \cite{tennakoon2021consensus}. Note: the algorithm of \cite{tennakoon2021consensus} is using uniform sampling at fixed Hamming level (so is implicitly providing a Hamming measure based result), though there it was interpreted as a biased way to estimate uniform measure influence, rather than, as here, a new measure of influence. \cite{tennakoon2021consensus} empirically optimised their sampling level - hence we do not experiment with Hamming sampling here.

For comparison we include results from: RANSAC \cite{fischler1981random}, Lo-RANSAC \cite{chum2003locally}, $A^*$-NAPA-DIBP \cite{cai2019consensus} (We only use it to generate ground truth solutions for synthetic data with low rate of outliers.), MBF \cite{tennakoon2021consensus}, L1 \cite{olsson2010outlier}. 

Experiments employed Matlab R2020b on a computer with Intel(R) i7-8700K CPU and 32GB RAM.
\subsection{Robust linear regression}

\textbf{Analysis of parameters in weighted influences estimation:} Evaluation of the weighted influence \eqref{estweightinf} requires two parameters: sample probability $q$ and sample size $h$. Here, we study the effect of $q$ and $h$ using synthetic data on the $2$-dimensional line fitting problem. We generate $n=15$ points\footnote{We choose the number of data points very low since the exact weighted influences require exponential time to compute as $n$ increases.} around a straight line and randomly select $30\%$ points as outliers, which are perturbed with uniformly distributed noise in range $[-4,-0.1)\cap (0.1,4]$. The remaining points are perturbed with uniformly distributed noise in $[-0.1,0.1]$. We set the inliers threshold $\varepsilon$ as $0.1$ in all experiments throughout this subsection. 

\begin{figure}[t]
  \centering
   \includegraphics[width=0.5\linewidth]{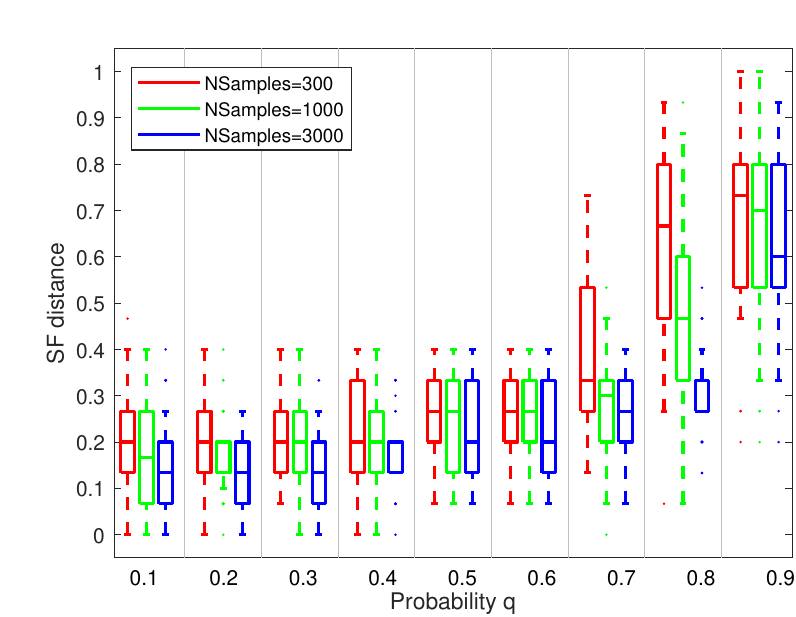}~
   \includegraphics[width=0.49\linewidth]{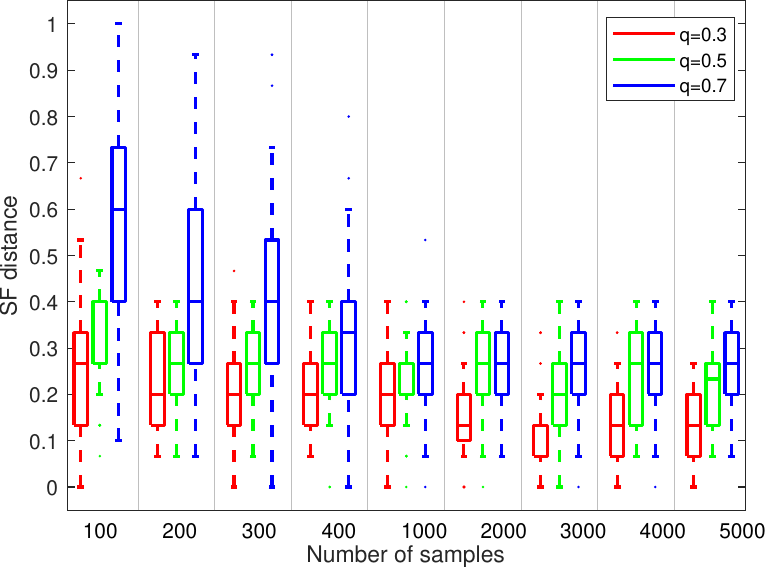}
   \caption{Comparison results for $2$-dimensional robust linear regression: Variation of the SF distance with respect to different (Left) sample probability $q$ and (Right) number of samples $h$. All experiments were run $50$ times.}
   \label{fig:linecompareqh}
   \vspace{-3mm}
\end{figure}

Let $\pmb{r}_{es}$ and $\pmb{r}_{ex}$ be the top $k$ (in decreasing order) of estimated weighted influences (calculated by \eqref{estweightinf}) and exact weighted influences (calculated by \eqref{weightinfdef}) of all points, respectively. The value $k$ is determined by the number of outliers (found by $A^*$-NAPA-DIBP). We measure the difference between $\pmb{r}_{es}$ and $\pmb{r}_{ex}$ by the normalized Spearman Footrule (SF) distance \cite{fagin2003comparing}. Denote a set of elements contained in $\pmb{r}_{\bullet}$ by $\mathcal{R}(\pmb{r}_{\bullet})$ and the position of the element $z\in \mathcal{R}(\pmb{r}_{\bullet})$ in $\pmb{r}_{\bullet}$ by $z^{\pmb{r}_{\bullet}}$. The normalized SF distance between $\pmb{r}_{es}$ and $\pmb{r}_{ex}$ is given by
\begin{align*}
F(\pmb{r}_{es},\pmb{r}_{ex}) = \frac{1}{k(k+1)}\sum_{z\in \mathcal{R}(\pmb{r}_{es})\cap \mathcal{R}(\pmb{r}_{ex})} \vert z^{\pmb{r}_{es}}-z^{\pmb{r}_{ex}} \vert,
\end{align*}
where we use $k+1$ for $z^{\pmb{r}_\bullet}$ if $z\not\in \mathcal{R}(\pmb{r}_\bullet)$. The smaller the SF distance is, the more likely that $\pmb{r}_{es}$ and $\pmb{r}_{ex}$ share the same top $k$ order of weighted influences.

For fixed number of samples ($h=300,1000,3000$), we vary the probability $q$ from $0.1$ to $0.9$; and for fixed probability ($q=0.3,0.5,0.7$), we vary the number of samples from $100$ to $5000$. The results are shown in Figure \ref{fig:linecompareqh}, from which we can see that the SF distance is relatively small when $q\in (0.2,0.4)$; and for more samples, the SF distance is smaller. In the following experiments, we will set the probability $q$ between $(p+1)/n$ and $0.4$ and sample size $h$ between $100$ to $500$ to maintain a trade-off between estimation accuracy and run time. Note that the SF distance is not able to distinguish points in $\pmb{r}_{es}$ or $\pmb{r}_{ex}$ who share the same weighted influences. These experiments also show the necessity of iteratively re-estimate weighted influences since evaluating all weighted influences once does not match the order of exact weighted influences perfectly.

\begin{figure}[t]
  \centering
   \includegraphics[width=0.5\linewidth]{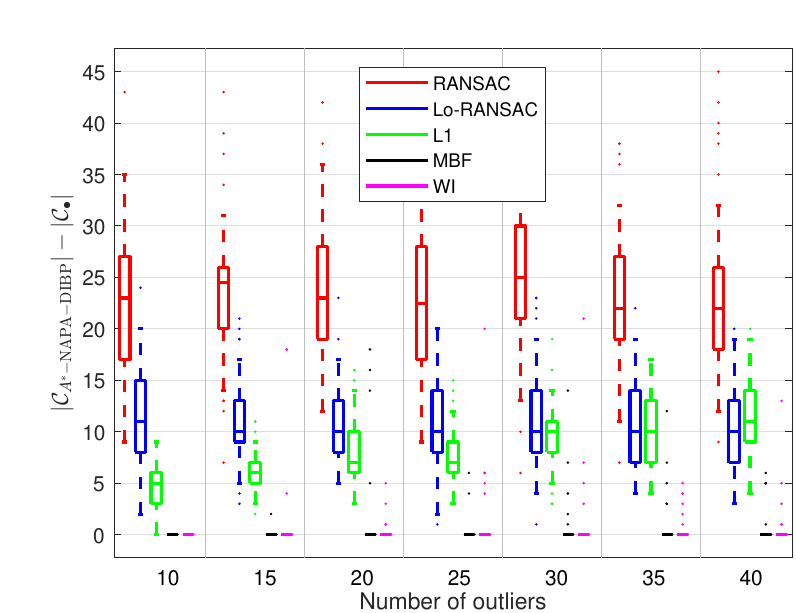}~
   \includegraphics[width=0.49\linewidth]{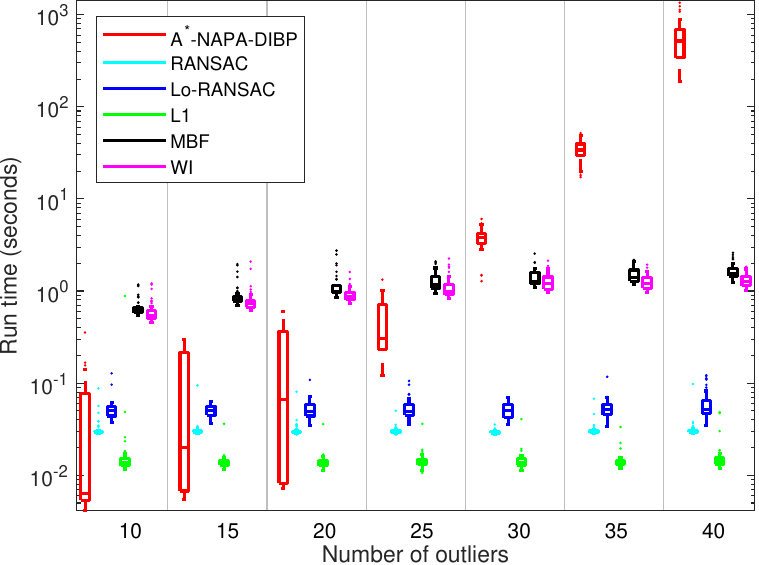}
   \caption{Comparison results for $8$-dimensional robust linear regression: Compare (Left) consensus error and (Right) run time with different number of outliers. The ground truth is found by $A^*$-NAPA-DIBP. All experiments were repeated $50$ times.}
   \label{fig:linecompareothers}
\end{figure}

\textbf{Relative Performance:} In this experiment, we consider $8$-dimensional linear regression problem, where the synthetic data are generated in the same way as we did in last experiments. The number of data points $n$ is chosen as $200$ and we vary the number of outliers from $10$ to $40$, which is limited by the computation time of $A^*$-NAPA-DIBP for generating ground truth. From Figure \ref{fig:linecompareothers}, we can find both Bernoulli weighted influences and the original MBF can achieve optimal solutions and the Bernoulli weighted method is faster than MBF in general. The effect of local expansion is compared in Appendix \ref{sec:local_expansion}.

\begin{table*}[t]
\caption{Results for linearised fundamental matrix estimation. Runtime of RANSAC and Lo-RANSAC is set to that of WI.}
\label{tab:fundmat}
\centering
\begin{tabular}{p{0.12\linewidth}p{0.1\linewidth} p{0.1\linewidth}p{0.1\linewidth}p{0.1\linewidth}p{0.1\linewidth}p{0.1\linewidth}}
\toprule
 &  & \pmb{104-108} & \pmb{198-201} & \pmb{417-420} & \pmb{579-582} & \pmb{738-742}\\
\midrule
\bf{L1} & Consensus & 252 & 264 & 317 & 497 & 429\\
& Time (s) & 0.85 & 0.07 & 0.04 & 0.06 & 0.09\\
\midrule
\bf{RANSAC} & Consensus & 266.55 & 284.00 & 353.30 & 497.05 & 432.60\\
& & (262,270) & (283,285) & (351,356) & (492,503) & (428,440)\\
\midrule
\bf{Lo-RANSAC} & Consensus & 268.55 & 284.90 & 355.25 & 501.45 & 439.35\\
& & (266,273) & (284,287) & (354,356) & (496,506) & (432,443) \\
\midrule
\bf{MBF} & Consensus & 271.70 & 287.70 & 358.95 & 507.15 & 444.20\\
& & (268,274) & (284,289) & (357,360) & (503,510) & (442,446) \\
 & Time (s) & 3.17 & 1.88 & 2.36 & 3.61 & 3.19\\
& & (2.65,4.41) & (1.65,2.47) & (1.96,2.66) & (3.17,4.36) & (2.68,3.56) \\
\midrule
\bf{WI} & Consensus & 271.55 & 288.35 & 359.05 & 507.65 & 444.15\\
& & (269,274) & (286,289) & (356,360) & (502,510) & (442,446)\\
&Time (s) & 2.64 & 1.75 & 2.10 & 3.42 & 3.09\\
& & (2.26,3.31) & (1.43,2.03) & (1.82,2.51) & (2.93,4.15) & (2.80,3.57) \\
\bottomrule
\end{tabular}
\end{table*}

\subsection{Linearised fundamental matrix estimation}

We further test the Bernoulli weighted influence  based approach on five image pairs from sequence ``00" of the KITTI Odometry dataset \cite{geiger2012we} in this subsection. For two given image pairs, SIFT keypoints \cite{lowe1999object} are detected and matched by the VLFeat toolbox \cite{vedaldi2010vlfeat}. Suppose $\pmb{p}_1,\pmb{p}_2$ are two features that are matched, where $\pmb{p}_i=(x_i,y_i,1)^T$ is the coordinate in view $i$, $i=1,2$, then each correspondence provides a linear constraint on the fundamental matrix $\pmb{F}\in\mathbb{R}^{3\times 3}$ as $\pmb{p}_1^T\pmb{F}\pmb{p}_2=0$. In this experiment, we consider the linearised version of fundamental matrix estimation \cite{hartley2000zisserman}, where the inliers threshold is set to $0.02$ for all image pairs. The iteration numbers of RANSAC and Lo-RANSAC are set to match the runtime of WI.

After running all algorithms $20$ times, the average consensus and runtime including their variance are reported in Table \ref{tab:fundmat}. We found that: (1) Although L1 is generally very fast, the  consensus size found is smaller than that of the Bernoulli based approach WI; (2) RANSAC and Lo-RANSAC are allowed to run the same time as WI, yet WI is still better in terms of returned consensus size; (3) WI achieves almost the same consensus size as MBF but with less time cost (on image pairs 198-201, 417-420, 579-582, WI finds a slightly higher average consensus size than MBF). Further detail is in Appendix \ref{sec:FM}.
\subsection{Homography estimation}

This experiment considers homography estimation together with linear residual model. For a set of correspondences from two views represented by $\{(\pmb{x}_i,\pmb{y}_i)\}$ with $\pmb{x}_i,\pmb{y}_i\in\mathbb{R}^2$, let $\tilde{\pmb{x}}_i=(\pmb{x}_i,1)^T$, $\tilde{\pmb{y}}_i=(\pmb{y}_i,1)^T$ be the homogeneous representation, homography estimation is to find a matrix $\pmb{H}\in\mathbb{R}^{3\times 3}$ such that $\tilde{\pmb{y}}_i=\pmb{H}\tilde{\pmb{x}}_i$ for inliers. Since $\pmb{H}$ is $8$ dimensional uniquely defined up to scale, it is possible to express homography estimation with linear residual as a linear regression problem (see Chapter 4 in \cite{hartley2000zisserman} for details). We use $4$ image pairs from the Zurich Buildings datasets (Building 5, 22, 28, 37). Similar to linearised fundamental matrix estimation, VLFeat toolbox is used for extracting SIFT features and correspondence matching. Note that each correspondence match produces two residual functions, which means the number of data points, inliers/outliers are doubled in optimisation.

\begin{figure}[ht]
  \centering
   \includegraphics[width=0.5\linewidth]{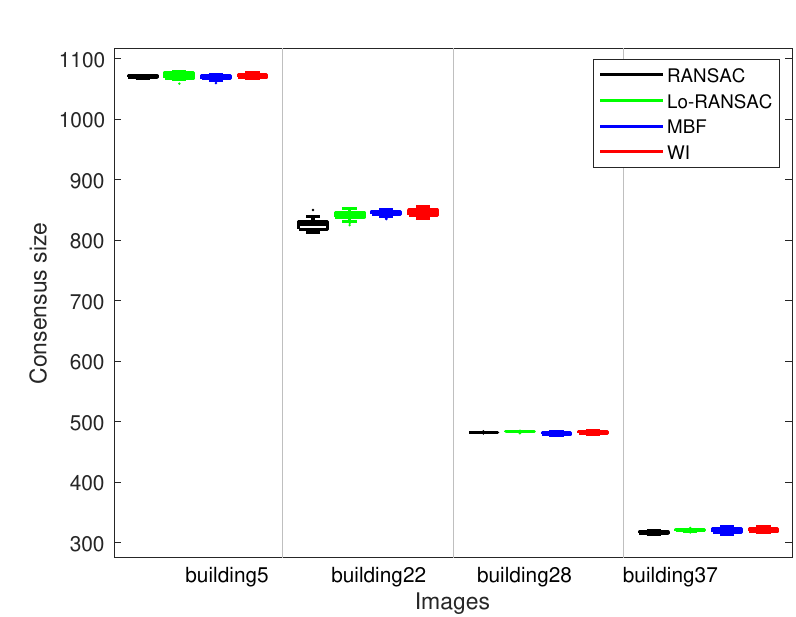}~
\includegraphics[width=0.47\linewidth]{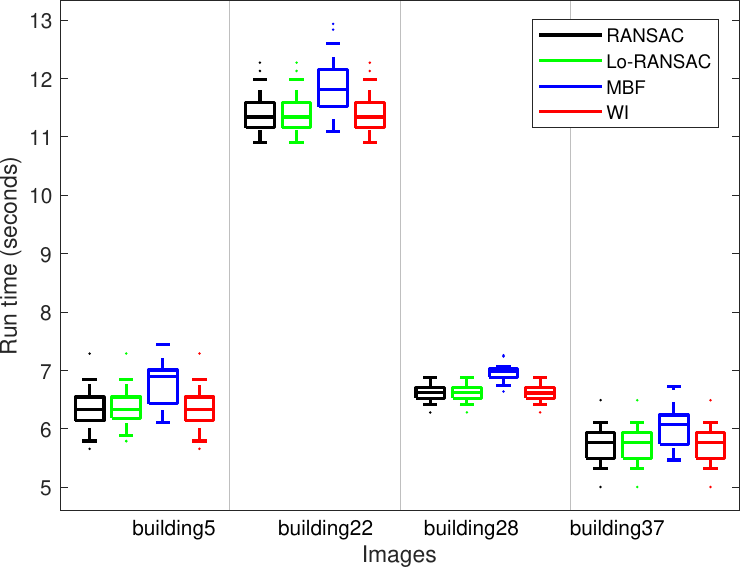}
   \caption{Comparison results for linearised homography estimation on $4$ Zurich Building image pairs: (Left) Compare consensus size and (Right) run time. All experiment were repeated $20$ times.}
   \label{fig:homocompareothers}
   \vspace{-3mm}
\end{figure}

In this experiment, we allow RANSAC and Lo-RANSAC to run for the same time as WI. The inliers threshold is set to $0.1$. From Figure \ref{fig:homocompareothers}, we find that WI can achieve (roughly) the best performance among the four sampling based methods. The run time of WI is smaller than that of MBF on average. Further comparison between WI and MBF can be found in Appendix \ref{sec:homog}.
\section{Conclusion}
 
We 
studied weighted influence in relation to the MaxCon problem. Our analysis is more general than \cite{tennakoon2021consensus} in terms of both the inclusion of weighted measures and in the study of  the ``non-ideal case''. Incorporating weighted influences into the algorithm of \cite{tennakoon2021consensus}, 
is straightforward. As demonstrated by experiments on both synthetic and real data, doing so is an effective alternative of \cite{tennakoon2021consensus} (roughly, saving some run time while achieving similar consensus or achieve better consensus with similar time budget). However, the main message is that, rather than viewing biased sampling as some way to (more efficiently) estimate uniform influence, and ``trusting'' that the bias (in the estimates) thereby introduced has no detrimental effect on the ordering of influences (specifically of outliers having larger influence than inliers), a proper study of biased measures allows us to characterise the changes in these influence measures and check for retention of the correct ordering. We address this for two ``natural'' families of biased measures: Bernoulli and Uniform at a fixed Hamming level.

 Of course our work has two obvious limitations: our  anaysis (of ideal cases) necessarily involves assumptions whose relevance to real world data could be challenged, and our experiments are not exhaustive. 

{\small
\bibliographystyle{ieee_fullname.bst}
\bibliography{egbib}
}

\appendix

\section{Proofs of theoretical results}
\label{sec:proofs}

\begin{customthm}{2.1}
If $f:\{0,1\}^n\rightarrow \{0,1\}$ is a monotone Boolean function, then ${\rm{Inf}}_i^q[f]=-\frac{1}{\sqrt{q(1-q)}}\hat{f}^q(\{i\})$.
\end{customthm}

\begin{proof}
The {\em $i$-th derivative operator} $D_i$ maps a Boolean function $f$ to a function $D_if$ defined by \cite{o2014analysis}
\begin{align}\label{defdo}
D_if(\pmb{b}):=f(\pmb{b}^{i\mapsto 1})-f(\pmb{b}^{i\mapsto 0}),
\end{align}
where $\pmb{b}^{i\mapsto a}=(b_1,\cdots,b_{i-1},a,b_{i+1},\cdots,b_n)$. Since $f:\{0,1\}^n\rightarrow \{0,1\}$ is Boolean-valued, we have
\begin{align}
D_if(\pmb{b})=\begin{cases}
\pm 1,~~&\mathrm{if}~f(\pmb{b})\neq f(\pmb{b}^{\oplus i}),\\
0, &\mathrm{if}~f(\pmb{b})=f(\pmb{b}^{\oplus i}),
\end{cases}
\end{align}
which means
\begin{align}\label{infder1}
{\rm{Inf}}^q_i[f]=\langle D_if,D_if\rangle_q.
\end{align}

By definition,
\begin{align*}
D_i\chi_S^q(\pmb{b})=\begin{cases}
-\frac{1}{\sqrt{q(1-q)}}\chi_{S\setminus \{i\}}^q(\pmb{b}),~~&\mathrm{if}~i\in S,\\
0,&\mathrm{if}~i\notin S.
\end{cases}
\end{align*}
Then 
\begin{align*}
D_if&=D_i\left(\sum_{S\subseteq [n]}\hat{f}^q(S)\chi_S^q\right)\\
&=-\frac{1}{\sqrt{q(1-q)}}\sum_{S:i\in S}\hat{f}^q(S)\chi_{S\setminus \{i\}}^q.
\end{align*}

If $f$ is monotone,
\begin{align*}
{\rm{Inf}}_i^q[f]=\langle D_if,\chi_{\emptyset}^q\rangle=-\frac{1}{\sqrt{q(1-q)}}\hat{f}^q(\{i\}).
\end{align*}
\end{proof}


\begin{customthm}{3.1}[Ideal Single Structure]\label{idealsingle}
If $n_0=1$, namely, $f$ is ideal with respect to a single maximum upper zero $\pmb{b}^{k_1}$, then, for $i\in S_{\pmb{b}^{k_1}}^1$ (inliers),
\begin{small}
\begin{equation*}
\label{eqn:inlier_ideal}
{\rm{Inf}}_i^q[f]=
(C^{n-1}_p-C^{{k_1}-1}_p)q^p(1-q)^{n-p-1},
\end{equation*}
\end{small}
and for $i\in S_{\pmb{b}^{k_1}}^0$ (outliers),
\begin{small}
\begin{equation*}
\label{eqn:outlier_ideal}
{\rm{Inf}}_i^q[f]=C^{n-1}_pq^p(1-q)^{n-p-1}+\sum_{l=p+1}^{k_1}C^{k_1}_lq^l(1-q)^{n-l-1},
\end{equation*}
\end{small}
which implies ${\rm{Inf}}_{i_1}^q[f]-{\rm{Inf}}_{i_2}^q[f]=C^{{k_1}-1}_pq^p(1-q)^{n-p-1}+\sum_{l=p+1}^{k_1}C^{k_1}_lq^l(1-q)^{n-l-1}>0$, where $i_1\in S_{\pmb{b}^{k_1}}^0$, $i_2\in S_{\pmb{b}^{k_1}}^1$.
\end{customthm}
 
\begin{proof}
For any $i\in S_{\pmb{b}^{k_1}}^1$, if $\pmb{b}\in L_{\leqslant p-1}$ or $L_{\geqslant p+2}$, then $f(\pmb{b})\equiv f(\pmb{b}^{\oplus i})$; if $\pmb{b}\in L_p$, then $f(\pmb{b})\neq f(\pmb{b}^{\oplus i})$ holds for $(C^{n-1}_p-C^{{k_1}-1}_p)$-possible vertices $\pmb{b}$; if $\pmb{b}\in L_{p+1}$, then $f(\pmb{b})\neq f(\pmb{b}^{\oplus i})$ holds also for $(C^{n-1}_p-C^{{k_1}-1}_p)$-possible vertices $\pmb{b}$. Therefore, we have
\begin{align*}
{\rm{Inf}}_i^q[f]&=(C^{n-1}_p-C^{{k_1}-1}_p)q^p(1-q)^{n-p}\\
&+(C^{n-1}_p-C^{{k_1}-1}_p)q^{p+1}(1-q)^{n-p-1}\\
&=(C^{n-1}_p-C^{{k_1}-1}_p)q^p(1-q)^{n-p-1}.
\end{align*}

For any $i\in S_{\pmb{b}^{k_1}}^0$, if $\pmb{b}\in L_{\leqslant p-1}$ or $L_{\geqslant k+2}$, then $f(\pmb{b})\equiv f(\pmb{b}^{\oplus i})$; if $\pmb{b}\in L_p$, then $f(\pmb{b})\neq f(\pmb{b}^{\oplus i})$ holds for $C^{n-1}_p$-possible vertices $\pmb{b}$; if $\pmb{b}\in L_{p+1}$, then $f(\pmb{b})\neq f(\pmb{b}^{\oplus i})$ holds for $(C^{n-1}_p+C^{{k_1}}_{p+1})$-possible vertices $\pmb{b}$; if $\pmb{b}\in L_l$ $(p+2\leqslant l\leqslant {k_1})$, then $f(\pmb{b})\neq f(\pmb{b}^{\oplus i})$ holds for $(C^{{k_1}}_{l-1}+C^{{k_1}}_{l})$-possible vertices $\pmb{b}$; if $\pmb{b}\in L_{{k_1}+1}$, then $f(\pmb{b})\neq f(\pmb{b}^{\oplus i})$ holds for $C^{{k_1}}_{{k_1}}$-possible vertices $\pmb{b}$. Therefore, we have
\begin{small}
\begin{align*}
{\rm{Inf}}_i^q[f]&=C^{n-1}_pq^p(1-q)^{n-p}+(C^{n-1}_p+C^{{k_1}}_{p+1})q^{p+1}\times \\
&(1-q)^{n-p-1}+\sum_{l=p+2}^{k_1}(C^{{k_1}}_{l-1}+C^{{k_1}}_{l})q^{l}(1-q)^{n-l}\\
&+C^{{k_1}}_{k_1}q^{{k_1}+1}(1-q)^{n-{k_1}-1}\\
&=C^{n-1}_pq^p(1-q)^{n-p-1}+\sum_{l=p+1}^{k_1}C^{k_1}_lq^l(1-q)^{n-l-1}.
\end{align*}
\end{small}
\end{proof}


\begin{customthm}{3.2}[Ideal Multi-Structure]\label{idealkthm}
Suppose $n_0>1$, \ie, $f$ is ideal with several upper zeros, then, $\forall~i\in \cap_{l=1}^{n_0}S_{\pmb{b}^{k_i}}^{i_l}$ (if non-empty), 
\begin{small}
\begin{equation}\label{idealk}
\begin{aligned}
{\rm{Inf}}_i^q[f]~&=~(C^{n-1}_p-\sum_{i_s=1}C^{k_s-1}_p)q^p(1-q)^{n-p-1}\\
&+\sum_{i_s=0}\sum_{l=p+1}^{k_s}C^{k_s}_lq^l(1-q)^{n-l-1}.
\end{aligned}
\end{equation}
\end{small}
\end{customthm}

\begin{proof}
We prove this theorem by induction on $n_0$. By Theorem \ref{idealsingle}, \eqref{idealk} is true for $n_0=1$. Suppose \eqref{idealk} holds for $n_0-1$, namely, $\forall~\hat{i}\in \cap_{l=1}^{n_0-1}S_{\pmb{b}^{k_i}}^{i_l}$,
\begin{small}
\begin{align*}
{\rm{Inf}}_{\hat{i}}^q[f]&=C^{n-1}_pq^p(1-q)^{n-p-1}\\
&+\sum_{\underset{1\leqslant s\leqslant n_0-1}{i_s=0}}\sum_{l=p+1}^{k_s}C^{k_s}_lq^l(1-q)^{n-l-1}\\
&-\sum_{\underset{1\leqslant s\leqslant n_0-1}{i_s=1}}C^{k_s-1}_pq^p(1-q)^{n-p-1}.
\end{align*}
\end{small}

Now we only have to prove that, $\forall~i\in \cap_{l=1}^{n_0}S_{\pmb{b}^{k_i}}^{i_l}$,
\begin{align*}
{\rm{Inf}}_i^q[f]&={\rm{Inf}}_{\hat{i}}^q[f]\\
&+\begin{cases}
-C^{k_{n_0}-1}_pq^p(1-q)^{n-p-1},~~~&i_{n_0}=1,\\
\sum_{l=p+1}^{k_{n_0}}C^{k_{n_0}}_lq^l(1-q)^{n-l-1},&i_{n_0}=0.
\end{cases}
\end{align*}
When adding one more upper zero $\pmb{b}^{k_{n_0}}$, for any $i\in \cap_{l=1}^{n_0-1}S_{\pmb{b}^{k_i}}^{i_l}\cap S_{\pmb{b}^{k_{n_0}}}^1$, $i$-boundary edges will decrease by $C^{k_{n_0}-1}_p$ at level $p$ and $p+1$. That is, if $\pmb{b}\in L_{p}$ or $L_{p+1}$, then $f(\pmb{b})=f(\pmb{b}^{\oplus i})$ holds for $C^{k_{n_0}-1}_p$ possible vertices $\pmb{b}$. Then, the decrease amount for ${\rm{Inf}}_{\hat{i}}^q[f]$ is
\begin{align*}
&C^{k_{n_0}-1}_pq^p(1-q)^{n-p}+C^{k_{n_0}-1}_pq^{p+1}(1-q)^{n-p-1}\\
&=C^{k_{n_0}-1}_pq^p(1-q)^{n-p-1}.
\end{align*}
For any $i\in \cap_{l=1}^{n_0-1}S_{\pmb{b}^{k_i}}^{i_l}\cap S_{\pmb{b}^{k_{n_0}}}^0$, $i$-boundary edges will increase. In details, if $\pmb{b}\in L_{k_{n_0}+1}$, then $f(\pmb{b})\neq f(\pmb{b}^{\oplus i})$ holds for $C^{k_{n_0}}_{k_{n_0}}=1$ possible vertex $\pmb{b}$, if $\pmb{b}\in L_l$ $(p+2\leqslant l\leqslant k_{n_0})$, then $f(\pmb{b})\neq f(\pmb{b}^{\oplus i})$ holds for $(C^{k_{n_0}}_l+C^{k_{n_0}}_{l-1})$-possible vertices $\pmb{b}$, if $\pmb{b}\in L_{p+1}$, the possible vertices $\pmb{b}$ have $C^{k_{n_0}}_{p+1}$. Then, the increase amount for ${\rm{Inf}}_{\hat{i}}^q[f]$ is
\begin{small}
\begin{equation*}
\begin{aligned}
&C^{k_{n_0}}_{k_{n_0}}q^{k_{n_0}+1}(1-q)^{n-k_{n_0}-1}\\
&+\sum_{l=p+2}^{k_{n_0}}(C^{k_{n_0}}_l+C^{k_{n_0}}_{l-1})q^l(1-q)^{n-l}\\
&+C^{k_{n_0}}_{p+1}q^{p+1}(1-q)^{n-p-2}\\
&=\sum_{l=p+1}^{k_{n_0}}C^{k_{n_0}}_lq^l(1-q)^{n-l-1},
\end{aligned}
\end{equation*}
\end{small}
which complete this proof.
\end{proof}


\begin{customthm}{3.3}\label{nonidealthm}
If $f$ is non-ideal, then the weighted influence $\tilde{f}^q(S^{\pmb{i}})$ is given by
\begin{small}
\begin{equation}\label{nonidealeq}
\begin{aligned}
&(C^{n-1}_p-\sum_{\underset{1\leqslant s\leqslant n_0}{i_s=1}}C^{k_s-1}_p+\sum_{\underset{1\leqslant s\leqslant m_0}{i_{n_0+s}=0}}C^{\alpha_s-1}_p)q^p(1-q)^{n-p-1}\\
&+(\sum_{\underset{1\leqslant s\leqslant n_0}{i_s=0}}\sum_{l=p+1}^{k_s}C^{k_s}_l-\sum_{\underset{1\leqslant s\leqslant m_0}{i_{n_0+s}=1}}\sum_{l=p+1}^{\alpha_s}C^{\alpha_s}_l)q^l(1-q)^{n-l-1}.
\end{aligned}
\end{equation}
\end{small}
where $\tilde{f}^q(S^\bullet)$ doesn't exist if $S^\bullet=\emptyset$.
\end{customthm}

To better understand Theorem \ref{nonidealthm}, let us consider the simplest non-ideal case where $n_0=2$ and $m_0=1$.

\begin{customthm}{3.4}\label{nonideal21thm}
The existing influences are represented as
\begin{small}
\begin{align*}
\tilde{f}^q(S^{(110)})&=\left(C^{n-1}_p-C^{k_1-1}_p-C^{k_2-1}_p+C^{\alpha_1-1}_p\right)\times \\
&q^p(1-q)^{n-p-1},\\
\tilde{f}^q(S^{(101)})&=\left(C^{n-1}_p-C^{k_1-1}_p\right)q^p(1-q)^{n-p-1}\\
&+\left(\sum_{l=p+1}^{k_2}C^{k_2}_l-\sum_{l=p+1}^{\alpha_1}C^{\alpha_1}_l\right)q^l(1-q)^{n-l-1},\\
\tilde{f}^q(S^{(011)})&=\left(C^{n-1}_p-C^{k_2-1}_p\right)q^p(1-q)^{n-p-1}\\
&+\left(\sum_{l=p+1}^{k_1}C^{k_1}_l-\sum_{l=p+1}^{\alpha_1}C^{\alpha_1}_l\right)q^l(1-q)^{n-l-1},\\
\tilde{f}^q(S^{(001)})&=C^{n-1}_pq^p(1-q)^{n-p-1}\\
&+\left(\sum_{l=p+1}^{k_1}C^{k_1}_l+\sum_{l=p+1}^{k_1}C^{k_2}_l-\sum_{l=p+1}^{\alpha_1}C^{\alpha_1}_l\right)\times \\
&q^l(1-q)^{n-l-1},
\end{align*}
\end{small}
which implies 
\begin{small}
\begin{align*}
\tilde{f}^q(S^{(101)})<\tilde{f}^q(S^{(001)}),~~ \tilde{f}^q(S^{(011)})<\tilde{f}^q(S^{(001)}).
\end{align*}
\end{small}
\end{customthm}

\begin{proof}
By Theorem \ref{idealkthm}, we only have to prove
\begin{align*}
\tilde{f}^q(S^{(\bullet\bullet i_3)})&=\tilde{f}^q(S^{(\bullet\bullet}))\\
&+\begin{cases}
C^{\alpha_1-1}_pq^p(1-q)^{n-p-1},~~~&i_3=0,\\
\displaystyle -\sum_{l=p+1}^{\alpha_1}C^{\alpha_1}_lq^l(1-q)^{n-l-1},&i_3=1.
\end{cases}
\end{align*}
$\forall~i\in S^{(\bullet\bullet 0)}$, the vertices $\pmb{b}\in L_p$ or $L_{p+1}$ for $f(\pmb{b})\neq f(\pmb{b}^{\oplus i})$ have increased $C^{\alpha_1-1}_p$ because of the ovelap sub-cube $B_{\pmb{b}^{\alpha_1}}$, then
\begin{align*}
\tilde{f}^q(S^{(\bullet\bullet 0)})=&\tilde{f}^q(S_{\bullet\bullet })+C^{\alpha_1-1}_pq^p(1-q)^{n-p}\\
&+C^{\alpha_1-1}_pq^{p+1}(1-q)^{n-p-1}\\
=&\tilde{f}^q(S^{(\bullet\bullet) })+C^{\alpha_1-1}_pq^p(1-q)^{n-p-1}.
\end{align*}

$\forall~i\in S^{(\bullet\bullet 1)}$, if $\pmb{b}\in L_{\alpha_1+1}$, then $i$-boundary edges decrease by $C^{\alpha_1}_{\alpha_1}$, if $\pmb{b}\in L_l$ $(p+2\leqslant l\leqslant \alpha_1)$, then $i$-boundary edges decrease by $C^{\alpha_1}_l+C^{\alpha_1}_{l-1}$, if $\pmb{b}\in L_{p+1}$, the decrease amount is $C^{\alpha_1}_{p+1}$. Therefore,
\begin{align*}
\tilde{f}^q(S^{(\bullet\bullet 1)})=&\tilde{f}^q(S^{(\bullet\bullet) })-(C^{\alpha_1}_{\alpha_1}q^{\alpha_1+1}(1-q)^{n-\alpha_1-1}\\
&+\sum_{l=p+2}^{\alpha_1}(C^{\alpha_1}_l+C^{\alpha_1}_{l-1})q^l(1-q)^{n-l}\\
&+C^{\alpha_1}_{p+1}q^{p+1}(1-q)^{n-p-1})\\
=&\tilde{f}^q(S^{(\bullet\bullet) })-\sum_{l=p+1}^{\alpha_1}C^{\alpha_1}_lq^l(1-q)^{n-l-1},
\end{align*}
which complete the proof.
\end{proof}

By Theorem \ref{idealkthm} and Theorem \ref{nonideal21thm}, Theorem \ref{nonidealthm} can be proved by induction on $m_0$.

\section{Uniform Hamming measure Influence}
\label{sec:hamming}

Consider the ideal single structure case
define in Theorem \ref{idealsingle}. 
Firstly, it is easy to see the for the ideal single structure, any algorithm that starts with a feasible set of size large than the combinatorial dimension, and then greedily adds points if the larger set remains feasible, will obtain the MaxCon solution. So it is an easy problem with obvious solution strategies. But if we did decide to use influences we can note that
from equation \eqref{eqn:inlier_ideal} the inlier influences came from counting feasible/infeasible transitions between levels $p$ and $p+1$ only. Thus for sampling uniform Hamming level above level $p+1$, there are no feasibility/infeasibility transitions caused by inliers. In other words, the Uniform Hamming influence measure, at that level or above, will be exactly zero for inliers. Since the influences of outliers, for the same measure will be non-zero, this seems to promise a remarkably efficient sampling strategy - one could eliminate outliers at the first sample that revealed a count for the associated influence - without the need to continue with the full estimation process.  Of course, practically, this is too good to hold for real data - it is very brittle to our strict assumptions here. Nonetheless, it does hint at the usefulness of a less brittle strategy of early termination of the counting process once a count reaches some degree of statistically significantly higher than the rest, rather than a set number of samples always being used: a future research topic. 

Now consider the ideal multiple structure setting. 
From equation \eqref{idealk},  the influence accrued by being inlier to some structure (first term) is only accrued between level $p$ and $p+1$. The subtraction is due to feasibility transitions ``that didn't occur'' because the subset with added inlier remains within the same structure). So once again, sampling above level $p$ will not ``see'' those counts. But since inliers to one structure are outliers to another (we assumed no significant overlap): hence the influence of inliers to any structure will not be zero - different to the single structure case, as all structures are outliers to (all) other structures and thus accrue influence from the second term in the equation. It is also easy to see that so long as the level is ``not too high'' (above the largest structure) the Hamming sampled influences will be an appropriate guide (influence of inliers of a larger structure will be smaller). (In that second term the largest structure is excluded from the sum over $i_s=0$, when calculating influence for that structure.) 

In more detail, since a point is a member of at most one structure (we forbid overlaps in the definition of ``ideal''), we observe that an inlier to any structure is associated with only one term in the subtraction and 

Analysis of the non-ideal case is complicated (hugely) by the complex combinatorics of possible overlaps. Nonetheless, for structures with little overlap with any other (we would argue the majority of structures in situations of interest) the ``perurbation'' from the ideal case calculations will be minimal. For situations with very large overlap in structures, one could alternatively view these as minor variants of one and the same structure (simply including a few extra points and losing one or two) and thus - with respect to the overlaps involving the largest structure, these could be considered as minor sub-optimal variants and essentially recovering one of the slightly smaller variants, compared with the actual optimal, is something of likely minor practical consequence. Of course, we realise that such observations are far short of conclusive argument and we make no claims of otherwise. 

\section{Algorithms based on Bernoulli weighted influences}
\label{sec:alg}

Algorithm \ref{alg1} is essentially similar to that presented in \cite{tennakoon2021consensus}, where $p+1$ is the combinatorial dimension of the prescribed model, the function $f$ is evaluated as
\begin{small}
\begin{align}
f(\mathcal{I}) := \mathbb{I}(\min_{\pmb{\theta}}\max_{\pmb{x}_i\in \mathcal{I}}r(\pmb{x}_i,\pmb{\theta})\leqslant \varepsilon)
\end{align}
\end{small}
with $\mathbb{I}(\cdot)$ the indicator function. The key difference is how we evaluate the estimated weighted influences $\widetilde{{\rm{Inf}}}_i^q[f]$.

\begin{algorithm}
\SetKwFunction{isOddNumber}{isOddNumber}
\SetKwInput{Input}{Input}
\SetKwInput{Output}{Output}

\KwIn{Dataset $\mathcal{X}=\{\pmb{x}_i\}_{i=1}^n$, probability $q\in (0,1)$, sample size $h$, threshold $\varepsilon>0$.}
\KwOut{Inliers set $\mathcal{I}^*$}

Initialization: $\mathcal{I}\leftarrow \pmb{1}_{1\times n}$.

\While{$|\mathcal{I}|>p$}{
Solve the minmax problem
$$
\min_{\pmb{\theta}}\max_{\pmb{x}_i\in\mathcal{I}} r(\pmb{x}_i,\pmb{\theta}),
$$ 
to get a basis $\mathcal{B}$.

Evaluate the estimated weighted influences $\widetilde{{\rm{Inf}}}_i^q[f]$ for $i\in \mathcal{B}$ by
\begin{small}
\begin{align*}
\widetilde{{\rm{Inf}}}_i^q[f]=-\frac{1}{h\sqrt{q(1-q)}}\sum_{j=1}^h f(\pmb{b}_j)q_{-}^{b_{j,i}}q_{+}^{1-b_{j,i}}\mu_q(\pmb{b}_j).
\end{align*}
\end{small}

$\mathcal{I}\leftarrow \mathcal{I}\setminus \arg\max_i\{\widetilde{{\rm{Inf}}}_i^q[f] ~|~ i\in \mathcal{B}\}$.

\If{$f(\mathcal{I})=0$}{
$\mathcal{I}^*\leftarrow \mathcal{I}$.

Break.
}

Conduct Algorithm \ref{alg2} for local expansion to add possible missing inliers.
}

\KwRet{$\mathcal{I}^*$}
\caption{Consensus maximisation using weighted influences (WI)}\label{alg1}
\end{algorithm}

\begin{algorithm}
\SetKwFunction{isOddNumber}{isOddNumber}
\SetKwInput{Input}{Input}
\SetKwInput{Output}{Output}

\KwIn{Dataset $\mathcal{X}=\{\pmb{x}_i\}_{i=1}^n$, threshold $\varepsilon>0$, initial solution $\mathcal{I}$.}
\KwOut{Inliers set $\mathcal{I}$}

$\rm{Candidates}\leftarrow \mathcal{X}\setminus \mathcal{I}$.

\For{$i$ in $\rm{Candidates}$}{

$\mathcal{I}\leftarrow \mathcal{I}\cup \{i\}$.

\If{$f(\mathcal{I})=1$}{
$\mathcal{I}\leftarrow \mathcal{I}\setminus \{i\}$.
}
}
\KwRet{$\mathcal{I}$}
\caption{Local expansion step}\label{alg2}
\end{algorithm}

\section{Comparison of the effect of local expansion in MBF and WI}
\label{sec:local_expansion}

In this section, we will compare the effect of local expansion in MBF and WI using the example of $8$-dimensional linear regression. The experiment setting is the same as Subsection 4.1 in the main paper. We denote MBF and WI without local expansion by MBF-nL and WI-nL, respectively.

\begin{figure}
  \centering
   \includegraphics[width=0.8\linewidth]{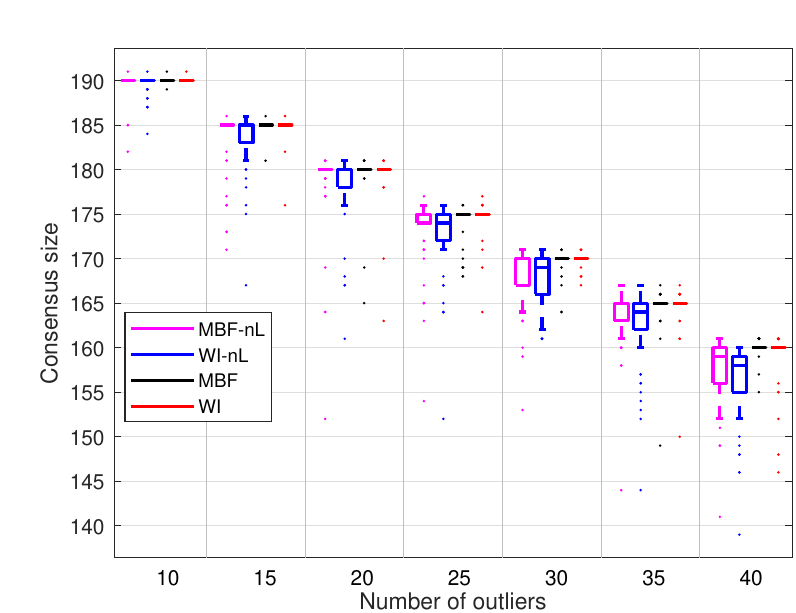}\\
    \vspace{3mm}
   \includegraphics[width=0.8\linewidth]{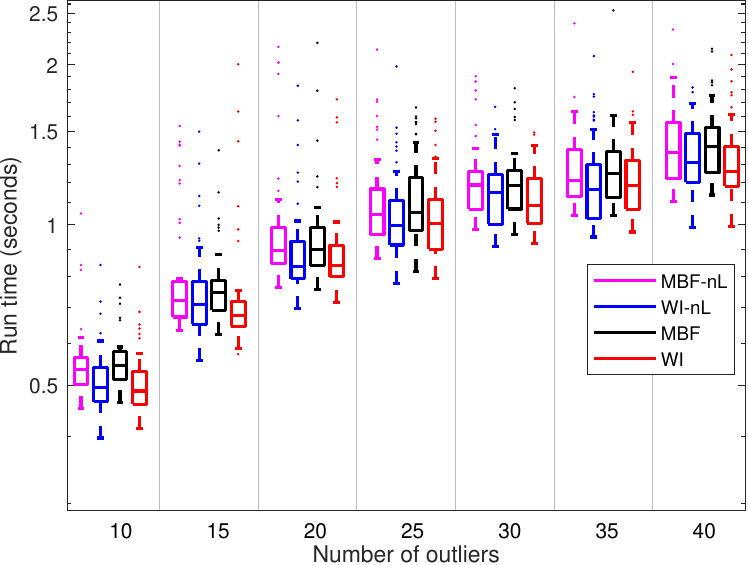}
   \caption{Comparison results of $8$ dimensional linear regression with synthetic data: (Top) consensus size and (Bottom) runtime. All experiments were repeated $50$ times.}
   \label{fig:compareLocal}
\end{figure}

From Figure \ref{fig:compareLocal}, we find that the number of inliers returned by the proposed method without local expansion WI-nL is less than that of MBF-nL, however, with the help of local expansion, both WI and MBF can find the same number of inliers. More importantly, our method WI (WI-nL) is generally faster than MBF (MBF-nL), especially in the presence of higher number of outliers. 

\section{Further results on linearised fundamental matrix estimation}
\label{sec:FM}

This section further examines the performance of our proposed method on linearised fundamental matrix estimation on the KITTI dataset that is used in the main paper. In this experiment, we choose the confidence $\rho=0.99$ for the standard stopping criteria in both RANSAC and Lo-RANSAC. The average consensus size and runtime including their variance over $20$ repeated runs are shown in Table \ref{tab:fundmatsm}.

From Table \ref{tab:fundmatsm}, we can find that: (1) With respect to the standard stopping criteria, RANSAC-p and Lo-RANSAC-p are much faster than the proposed method, however, they sacrifice the consensus size a lot, especially on the image pair 738-742; (2) With the help of local expansion, both MBF and WI improve the returned consensus size from MBF-nL and WI-nL with a small amount of extra time budget. Moreover, without local expansion, WI-nL is slightly better than MBF-nL in terms of returned consensus size and runtime, on average. 

\begin{figure}[t]
  \centering
   \includegraphics[width=0.5\linewidth]{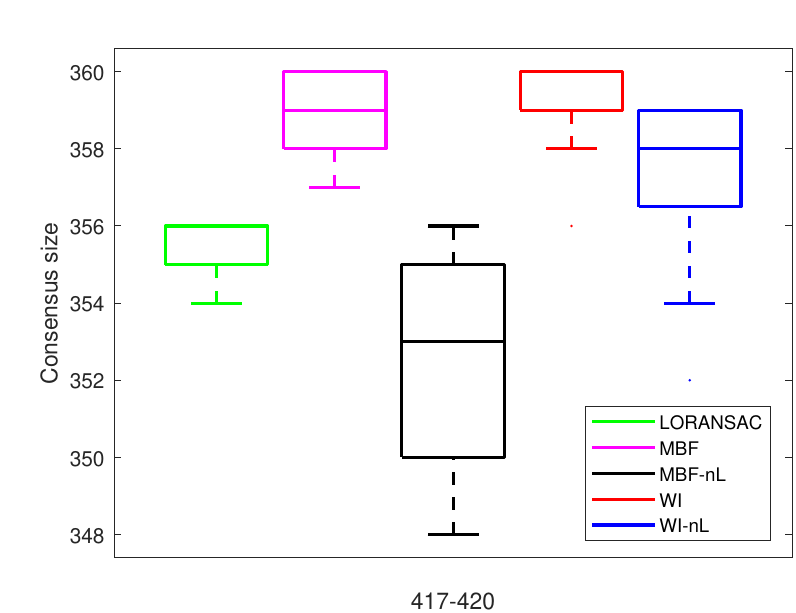}~
   \includegraphics[width=0.49\linewidth]{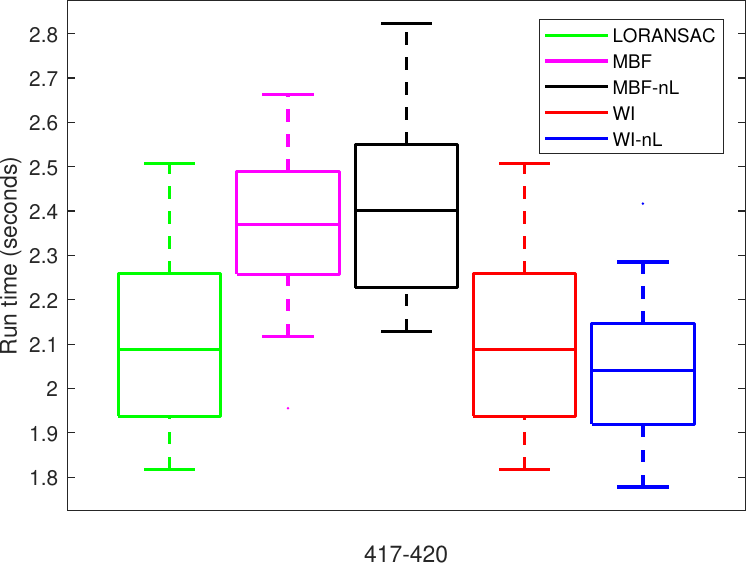}\\
   \includegraphics[width=0.5\linewidth]{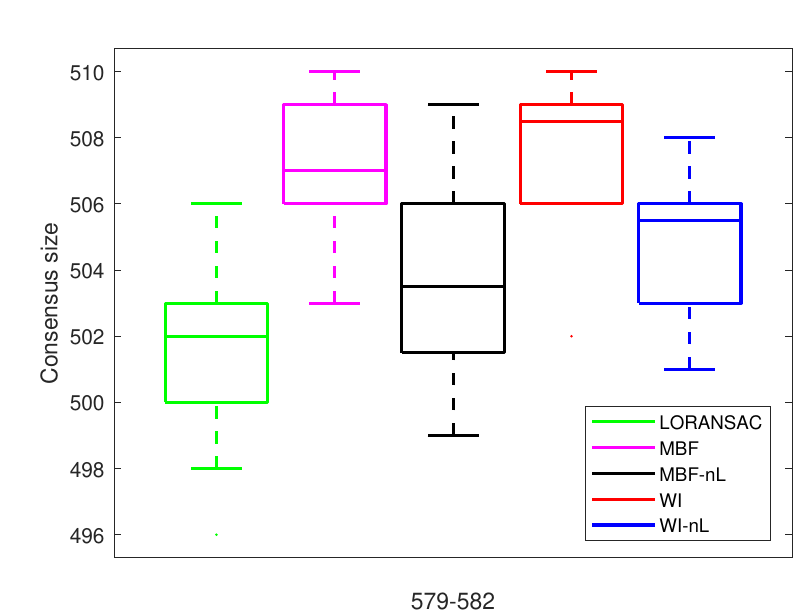}~
\includegraphics[width=0.49\linewidth]{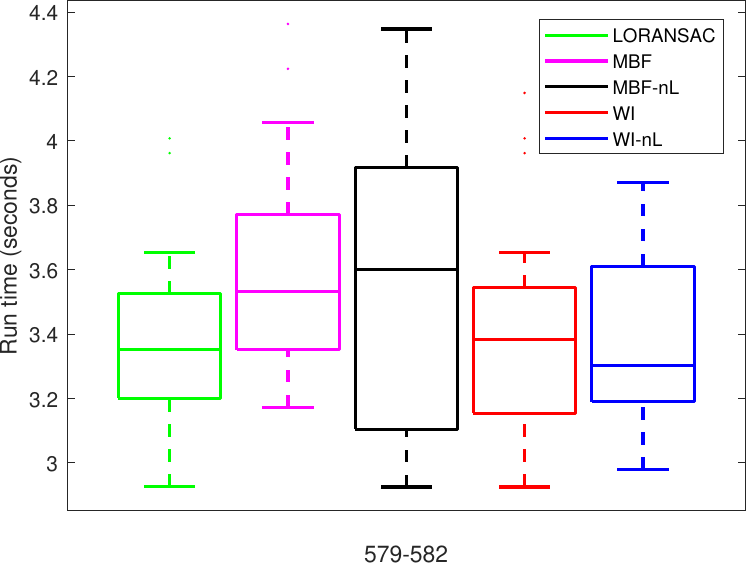}
   \caption{Distributions of consensus size (left column) and run time (right column) of linearised fundamental matrix estimation on KITTI image pairs $417-420$ (top row) and $579-582$ (bottom row).}
   \label{figsm:fundkitti34}
\end{figure}

To further compare the performance of Lo-RANSAC, MBF/MBF-nL and WI/WI-nL, we plot the distributions of consensus size and runtime on the image pair $417-420$ and $579-582$ in Figure \ref{figsm:fundkitti34}. From which, we can see that although WI and MBF can get similar average consensus size, WI has a higher probability to achieve better results with less time budget. Obviously, the more iterations of RANSAC and Lo-RANSAC use, the higher consensus size they return. However, our method as well as MBF can increase results by sampling more vertices in the Boolean cube to get more accurate (weighted) influences. 

A fair and safe conclusion is that on some datasets, WI and MBF (including their variants) perform better than RANSAC and Lo-RANSAC with some prescribed time (generally longer than the rule of thumb prescriptions for termination of those algorithms: thus when one is prepared to spend extra computation for better results, WI and MBF may be alternatives worth considering). More importantly, WI is able to achieve similar consensus size with less time budget, which means WI is an effective alternative of MBF. 

\begin{table*}[t]
\caption{Results for linearised fundamental matrix estimation. RANSAC-p and Lo-RANSAC-p refer to RANSAC and Lo-RANSAC with standard stopping criteria of the confidence $\rho=0.99$, respectively. MBF-nL and WI-nL refer to implementing MBF and WI without local expansion steps, respectively. All experiments were repeated over 20 random runs.}
\label{tab:fundmatsm}
\centering
\begin{tabular}{p{0.14\linewidth}p{0.09\linewidth} p{0.1\linewidth}p{0.1\linewidth}p{0.1\linewidth}p{0.1\linewidth}p{0.1\linewidth}}
\toprule
 &  & \pmb{104-108} & \pmb{198-201} & \pmb{417-420} & \pmb{579-582} & \pmb{738-742}\\
\midrule
\bf{RANSAC-p} & Consensus & 252.05 & 276.00 & 341.20 & 474.20 & 411.60\\
& & (238,266) & (267,282) & (317,351) & (453,496) & (401,425) \\
 & Time (s) & 0.01 & 0.01 & 0.01 & 0.01 & 0.01\\
& & (0.01,0.01) & (0.01,0.01) & (0.01,0.01) & (0.01,0.01) & (0.01,0.01) \\
\midrule
\bf{Lo-RANSAC-p} & Consensus & 264.15 & 281.75 & 354.05 & 492.25 & 423.15\\
& & (255,269) & (279,285) & (352,356) & (480,500) & (413,435) \\
 & Time (s) & 0.04 & 0.04 & 0.08 & 0.14 & 0.13\\
& & (0.01,0.07) & (0.01,0.07) & (0.03,0.13) & (0.08,0.27) & (0.05,0.26)\\
\midrule
\bf{MBF-nL} & Consensus & 261.80 & 285.60 & 352.75 & 503.75 & 441.40\\
& & (253,268) & (281,288) & (348,356) & (499,509) & (434,445)\\
& Time (s) & 3.03 & 1.84 & 2.41 & 3.57 & 3.25\\
& & (2.56,3.55) & (1.65,2.29) & (2.13,2.82) & (2.93,4.35) & (2.82,4.10)\\
\midrule
\bf{WI-nL} & Consensus & 269.50 & 287.40 & 357.40 & 504.75 & 441.90\\
& & (266,272) & (282,289) & (352,359) & (501,508) & (439,444)\\
&Time (s) & 2.53 & 1.67 & 2.07 & 3.40 & 3.02\\
& & (2.27,2.90) & (1.56,2.04) & (1.78,2.69) & (2.98,4.05) & (2.79,3.41)\\
\bottomrule
\end{tabular}
\end{table*}

\section{Further results on linearised homography estimation}
\label{sec:homog}

\begin{figure}[t]
  \centering
   \includegraphics[width=0.7\linewidth]{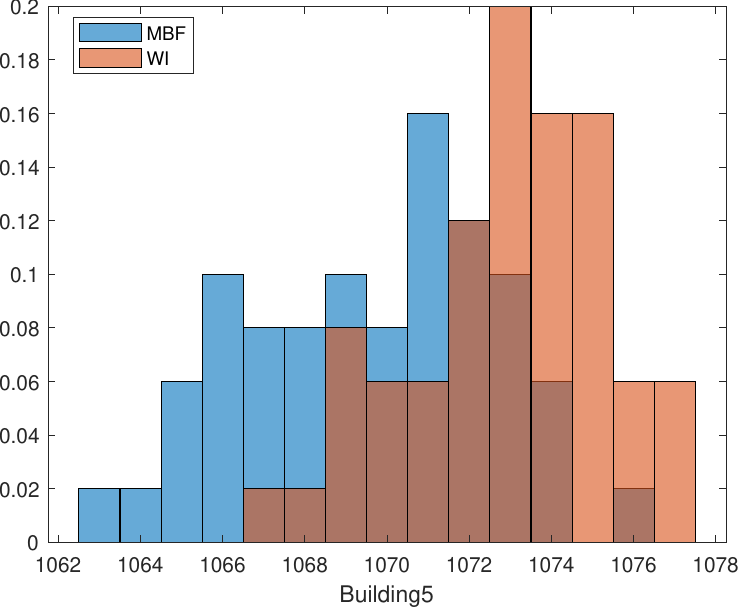}\\
    \vspace{3mm}
   \includegraphics[width=0.7\linewidth]{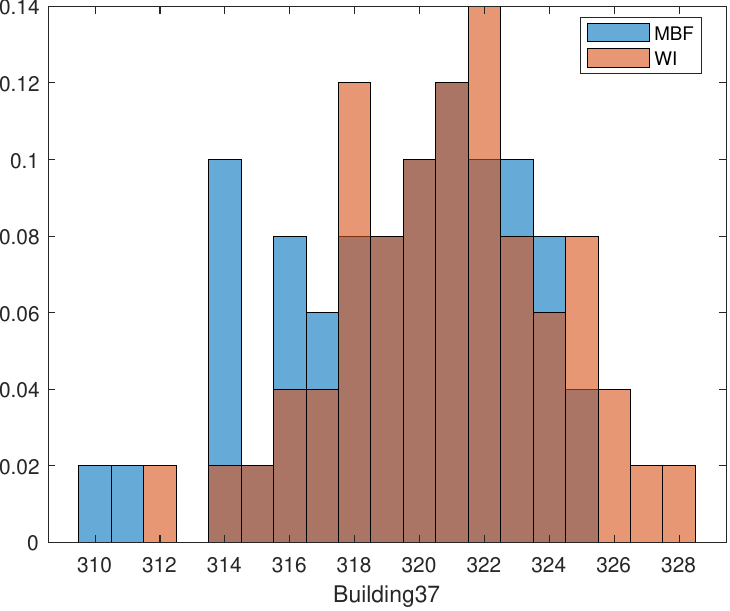}
   \caption{Distributions of consensus size returned by MBF and WI.}
   \label{fig:compareMBFWI}
\end{figure}

In this section, we compare the distributions of consensus size returned by MBF and WI, which is shown in Figure \ref{fig:compareMBFWI}. It can be seen that WI can achieve better results with high probability.

\end{document}